\documentclass[lettersize,journal]{IEEEtran}
\usepackage{amsmath,amsfonts}
\usepackage{algorithmic}
\usepackage{algorithm}
\usepackage{array}
\usepackage[caption=false,font=normalsize,labelfont=sf,textfont=sf]{subfig}
\usepackage{textcomp}
\usepackage{stfloats}
\usepackage{url}
\usepackage{verbatim}
\usepackage{graphicx}
\usepackage{cite}
\usepackage{makecell}
\usepackage{color}
\begin{document}

\title{How Simulation Helps Autonomous Driving: \\A Survey of Sim2real, Digital Twins, \\and Parallel Intelligence}
%\title{Sim2real, Digital Twins, and Parallel Intelligence in Autonomous Driving: A Survey}

\author{Xuemin Hu, Shen Li, Tingyu Huang, Bo Tang,~\IEEEmembership{Senior Member, IEEE}, Rouxing Huai, \\and Long Chen,~\IEEEmembership{Senior Member, IEEE}
        % <-this % stops a space
\thanks{This work was supported in part by the National Natural Science Foundation of China under Grant 62273135 and in part by the Natural Science Foundation of Hubei Province in China under Grant 2021CFB460. (corresponding author: Long Chen) }% <-this % stops a space
\thanks{Xuemin Hu, Shen Li, and Tingyu Huang are with the School of Artificial Intelligence, Hubei University, Wuhan, Hubei, 430062, China (e-mail: huxuemin2012@hubu.edu.cn, lishen\_stu@163.com, a1780585944@163.com)
	
Bo Tang is with the Department of Electrical and Computer Engineering, Worcester Polytechnic Institute, Worcester, MA, 01609, USA. (e-mail: btang1@wpi.edu)

Rouxing Huai is with Beijing Huairou Academy of Parallel Sensing, Beijing, 101499, China. (e-mail: xr.huai@qaii.ac.cn)

Long Chen is with the State Key Laboratory of Management and Control for Complex Systems, Institute of Automation, Chinese Academy of Sciences, Beijing, 100864, China, and also with the Waytous Inc. Beijing 100083, China. (e-mail: long.chen@ia.ac.cn)}}

\markboth{Journal of \LaTeX\ Class Files,~Vol.~XX, No.~XX, XXX~XXX}
{Shell \MakeLowercase{\textit{et al.}}: A Sample Article Using IEEEtran.cls for IEEE Journals}


\maketitle

\begin{abstract}
Safety and cost are two important concerns for the development of autonomous driving technologies. From the academic research to commercial applications of autonomous driving vehicles, sufficient simulation and real world testing are required. In general, a large scale of testing in simulation environment is conducted and then the learned driving knowledge is transferred to the real world, so how to adapt driving knowledge learned in simulation to reality becomes a critical issue. However, the virtual simulation world differs from the real world in many aspects such as lighting, textures, vehicle dynamics, and agents' behaviors, etc., which makes it difficult to bridge the gap between the virtual and real worlds. This gap is commonly referred to as the reality gap (RG). In recent years, researchers have explored various approaches to address the reality gap issue, which can be broadly classified into three categories: transferring knowledge from simulation to reality (sim2real), learning in digital twins (DTs), and learning by parallel intelligence (PI) technologies. In this paper, we consider the solutions through the sim2real, DTs, and PI technologies, and review important applications and innovations in the field of autonomous driving. Meanwhile, we show the state-of-the-arts from the views of algorithms, models, and simulators, and elaborate the development process from sim2real to DTs and PI. The presentation also illustrates the far-reaching effects and challenges in the development of sim2real, DTs, and PI in autonomous driving.
\end{abstract}


\section{Introduction}
\IEEEPARstart{A}{utonomous} driving, as an important part of intelligent transportation in the future, has great potential in alleviating traffic congestion and avoiding traffic accidents caused by human factors. In the past decades, researchers' efforts in this direction have increased, and many academic achievements have been made in autonomous driving \cite{hu2018dynamic, Zhang2021A, Part0, Part1, Part2}.

The application of autonomous driving in industry demands significant effort, especially for the planning and control tasks \cite{Hu2022Learning, Liu2020Finite, 9965619, 10122127}. It is necessary to consider not only the safety performance at the algorithm level, but also the cost at the real vehicle level, such as the price of high-precision sensors, radars, and cameras, as well as the collision damage of real vehicles, etc. The consequences of directly applying an immature algorithm in a real vehicle are immeasurable. Therefore, an experimental process is usually conducted in the high-fidelity simulator at first, and then deployed in the reality environment, which can greatly reduce the research and development costs. However, the simulation and the reality environment could be completely different, and there are always gaps between the simulated and real worlds such as lighting, textures, vehicle dynamics, and agents' behaviors, etc., which is usually named ``reality gap (RG)" \cite{kadian2020sim2real}.

\begin{figure}[tb!]
	\centering
	\includegraphics[width=3.5in]{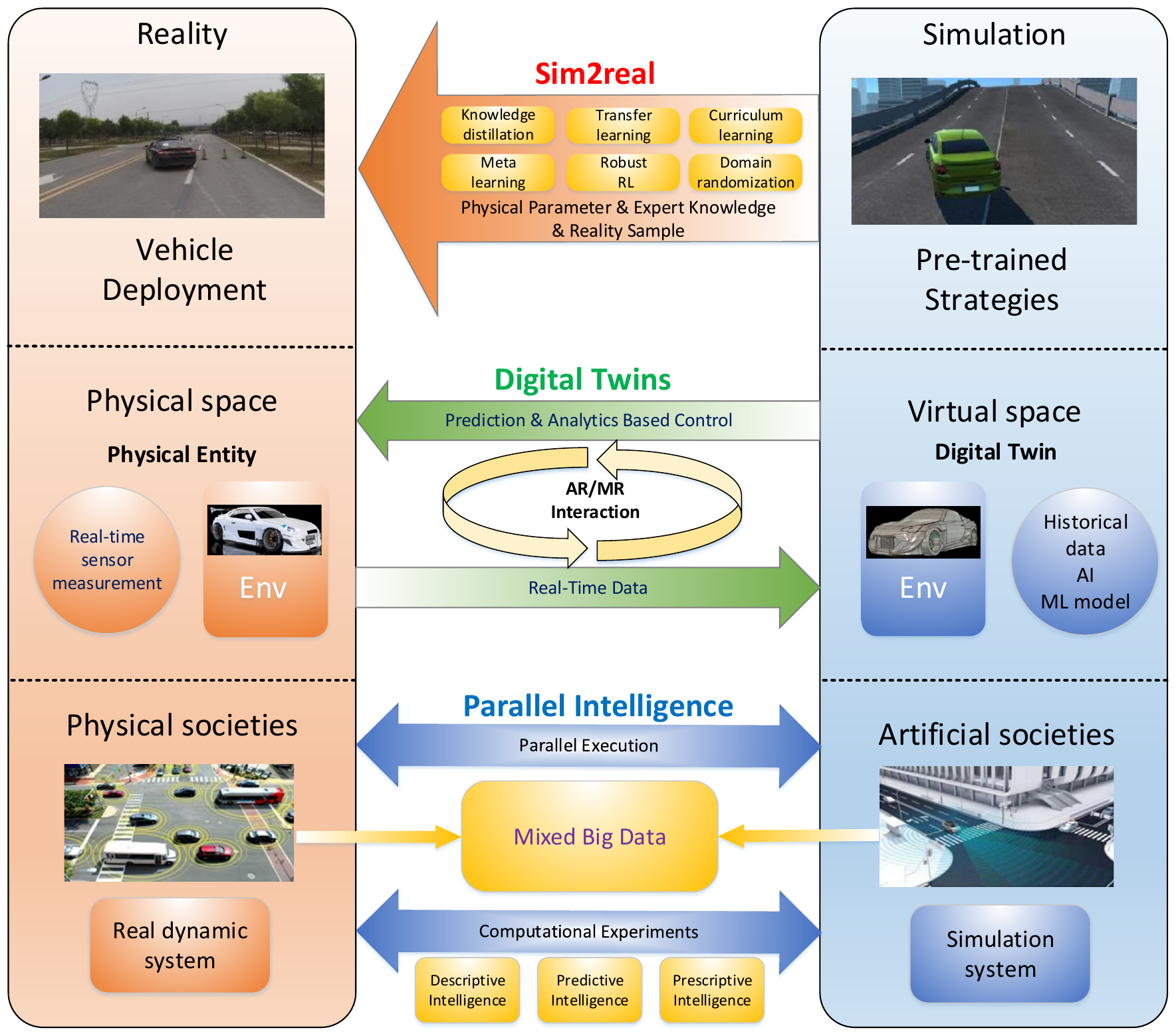}
	\caption{Leveraging computer simulations to improve real-world autonomous driving performance using three types of techniques: sim2real, digital twins (DTs), and parallel intelligence (PI). Sim2real mainly focus on improvement at the algorithm level. DTs are more concerned with scenario modeling and interaction. As a higher level technology, PI which has three-level functions including description, prediction, and prescriptive tends to construct mixed datasets and parallel computation of both artificial and physical scenarios.}
	\label{fig_1}
\end{figure}

In order to bridge the gap between the simulation and reality, many different methods have been proposed \cite{kang2019generalization}. At a high level, these methods are mainly divided into three categories: i) transferring knowledge from simulation to reality, ii) learning in digital twins, and iii) learning by parallel intelligence technologies, which are shown in Fig. \ref{fig_1}.

Sim2real refers to the process of transferring the strategies and knowledge learned from the simulated world to the real word to bridge the RG. In the field of autonomous driving, the core idea of sim2real is to train autonomous driving systems in simulation environment and then to apply them in real-world vehicles. In order to address the RG caused by the factors such as uneven sampling in the real world, too many physical parameters, insufficient expert experience, and imperfect dynamics models, etc., sim2real has gradually been developed using six kinds of methods, including curriculum learning, meta-learning, knowledge distillation, robust reinforcement learning, domain randomization, and transfer learning. The top part of Fig. \ref{fig_1} shows these methods and their relationships. Each method has its unique way to deal with the RG problem. For example, randomization of environmental parameters is proposed in domain randomization so that the parameters in the simulated world can cover those in the real world, making the strategies and knowledge learning from the simulated world applicable in the real world. However, the computational cost of sim2real is still a challenge, especially when dealing with complex and dynamic environment. The computational requirements for simulated complex environments are difficult to meet, which limits the scalability of sim2real methods.

Unlike sim2real methods, digital twin (DT)-based methods aim to construct a mapping of real-world physical entities in a simulation environment using the data from sensors and physical models to achieve the role of reflecting the entire lifecycle process of corresponding physical entities \cite{Edington2023A}. In autonomous driving, DTs are typically used for multi-scale modeling of the environment and vehicles. As shown in the middle part of Fig. \ref{fig_1}, data from real sensors are used to achieve motion planning of the twin body and physical vehicle in the virtual scenes through data interaction between the driving data analysis model and virtual reality. Some additional expert experience guides the modeling process to further narrow the RG. Moreover, the DT system is continuously learned and optimized through real-time data updates and interactions between the virtual and real twins to keep improving accuracy of the model. To achieve better virtual simulation effects, researchers combine virtual reality technologies with DTs. Argument reality (AR) and mixed reality (MR) technologies can provide better interactivity and visualization for digital twins. Users can interact and control the DT model using AR and MR technologies to achieve better physical simulation, visualization, and manipulation. In addition, AR and MR technologies can enhance DTs through real-time object tracking and virtualization, resulting in better interactive experiences and more advanced interactive applications.

Sim2real-based methods play a crucial role in adapting autonomous driving vehicles to the complexities of the real world by transferring learned knowledge from simulated environments, while DT-based methods allow autonomous driving vehicles to learn knowledge by synchronizing data from both the real and simulated worlds. In recent years, some researchers utilize parallel intelligence (PI) \cite{wang2016acp} to reduce RG. As a more advanced technology, PI combines the advantages of both sim2real and DT methods and can achieve better management and control for complex systems. Unlike DT, which only have a descriptive function, PI has three functions including description intelligence, prediction intelligence, and prescriptive intelligence, which are shown in the bottom part of Fig. \ref{fig_1}. In parallel intelligence techniques, researchers typically construct an artificial system which is mapped to a physical system to learn knowledge and gives feedback to the physical system. The artificial system learns knowledge from the data collected from the two systems through computational experiments for evaluation, and then the knowledge is applied to the real system through parallel execution with high real-time virtual-real interaction and online feedback.

In these three methods, what researchers need to do is to construct a virtual world.  Therefore, some researchers devote themselves to developing simulators of autonomous driving and robotics, such as AirSim\cite{shah2018airsim}, CARLA\cite{dosovitskiy2017carla}, etc. The mismatch between real and simulated settings can be minimized by providing training data and experience in these simulators, and robot agents can be deployed to the real world through sim2real methods.

This study surveys methods, applications, and development of bridging the reality gap in autonomous driving. To the best of our knowledge, this is the first survey to focus on dealing the RG from the perspectives of sim2real, digital twins, and parallel intelligence. In conclusion, our contributions are summarized as following three aspects.

\begin{itemize}
  \item A taxonomy of the literature is presented from the perspectives of sim2real, digital twins, and parallel intelligence, where particularly DT and PI methods are reviewed to address the reality gap issue for the first time.
  \item Methods and applications of handling the reality gap between simulation and reality in autonomous driving are comprehensively reviewed in this paper.
  \item Discussions of key challenges and opportunities are presented in this paper, offering insights of developing new sim2real, DT, and PI methods in autonomous driving.
\end{itemize}

The remainder of this paper is organized as follows. In the Section II, we introduce the methodologies and applications of sim2real. Technologies and applications of digital twins as well as the guiding technologies, AR and MR, in autonomous driving are introduced in Section III. In Section IV, parallel intelligence technologies for autonomous driving are presented. In Section V, we introduce the simulators that are used to implement the above related technologies. The future works and challenges are summarized in Section VI, and the conclusions are drawn in the final section.

% third para
\section{Simulation to reality transfer}
Autonomous driving algorithms require extensive testing before commercial applications. For safety and cost consideration, most of the current researches on new algorithms, especially for reinforcement learning-based methods, have focused on simulation environment \cite{lillicrap2015continuous,mnih2016asynchronous}. However, vehicle agents are trained to near-human levels in the simulation environment, their performance in the real world is still not comparable to that in the simulation environment. To overcome this issue, researchers have turned to sim2real transfer technologies, where the models are trained well in the simulation and then transferred to the real world. In this section, we present related approaches and technologies to achieve this goal, including curriculum learning, meta-learning, knowledge distillation, robust reinforcement learning, domain randomization, and transfer learning. Tab.1 presents some typical sim2real models which are reviewed in this section.

\begin{table*}[!tbp]%调节图片位置，h：浮动；t：顶部；b:底部；p：当前位置
	\caption{TYPICAL SIM2REAL MODELS IN AUTONOMOUS DRIVING.}
	\label{tab:1}
	\resizebox{\linewidth}{!}{
		\begin{tabular}{cccccc}%表格中的数据居中，c的个数为表格的列数
			\hline\hline
			\noalign{\smallskip}	
			Article & Category & Dataset/Simulator & Implemented Tasks & Description & Year\\
			\noalign{\smallskip}\hline\noalign{\smallskip}
			Qiao et al.\cite{qiao2018automatically} & \makecell{Curriculum \\learning} & NGSIM & \makecell {Driving planning \\decisions at crossroads} & \makecell{Use the automatic curriculum\\ generation method, which reduces\\ training time and provides better results.} &2018 \\
			
			\noalign{\smallskip}
			Bae et al.\cite{bae2021curriculum} & \makecell{Curriculum \\learning} & CarMaker & \makecell {Estimation of side camber \\and side slip angle} & \makecell{A curriculum strategy with task-specific scoring\\ and pacing functions is proposed to\\ reduce the side-slip angle side-tilt angle.} &2021 \\
			
			\noalign{\smallskip}
			Song et al.\cite{song2021autonomousovertaking} & \makecell{Curriculum \\learning} & \makecell{Gran Turismo \\Sport} & \makecell {Overtaking a car} & \makecell{A high-speed automatic overtaking system \\based on three-stage curriculum learning.} &2021 \\
			
			\noalign{\smallskip}
			Anzalone et al.\cite{anzalone2022end} & \makecell{Curriculum \\learning} & \makecell{MuJoCo} & \makecell {End-to-end competitive \\driving strategy} & \makecell{An end-to-end autonomous driving\\ reinforcement learning process\\ with five stages.} &2022 \\
			
			\noalign{\smallskip}
			Nagabandi et al.\cite{nagabandi2018learning} & \makecell{Meta-\\learning} & \makecell{MuJoCo} & \makecell {Simulation of dynamical\\ models online \\adaption of real scenarios} & \makecell{An online adaptive learning method \\for high-capacity dynamic models \\ to solve the simulation to reality problem.} &2018 \\
			
			\noalign{\smallskip}
			Jaafra et al.\cite{jaafra2019context} & \makecell{Meta-\\learning} & \makecell{CARLA} & \makecell {Autonomous driving strategy} & \makecell{A meta reinforcement learning approach to\\ embedding adaptive neural network controller\\s on top of adaptive meta-learning.} &2019 \\
			
			\noalign{\smallskip}
			Kar et al.\cite{kar2019metasim} & \makecell{Meta-\\learning} & \makecell{KITTI} & \makecell {Autonomous driving scene \\generation and rendering} & \makecell{Meta-Sim environment, where images \\and their corresponding realistic\\ ground images are acquired\\ through a graphics engine.} &2019 \\
			
			\noalign{\smallskip}
			Saputra et al.\cite{saputra2019distilling} & \makecell{Knowledge \\distillation} & \makecell{KITTI} & \makecell {Autonomous driving\\ trajectory prediction} & \makecell{Learning teacher's intermediate \\representations through \\attentional imitation loss and \\attentional cue training methods.} &2019 \\
			
			\noalign{\smallskip}
			Zhao et al.\cite{zhao2021sam} & \makecell{Knowledge \\distillation} & \makecell{CARLA} & \makecell {Driving\\ decision making} & \makecell{Let the student in the model try\\ to imitate the potential space of\\ the teacher in the base teacher model.} &2019 \\
			
			\noalign{\smallskip}
			Zhang et al.\cite{zhang2022pointdistiller} & \makecell{Knowledge \\distillation} & \makecell{KITTI} & \makecell {Point cloud map\\ feature extraction} & \makecell{A knowledge distillation method\\ based on point cloud map.} &2022 \\
			
			\noalign{\smallskip}
			Sautier et al.\cite{sautier2022image} & \makecell{Knowledge \\distillation} & \makecell{KITTI} & \makecell {3D image generation for\\ multimodal autonomous driving} & \makecell{A self-supervised knowledge \\distillation method.} &2022 \\
			
			\noalign{\smallskip}
			Li et al.\cite{li2022selfDistillation} & \makecell{Knowledge \\distillation} & \makecell{KITTI} & \makecell {Semantic segmentation of auto-\\nomous driving radar data} & \makecell{Transformer-based voxel feature \\encoder for robust\\ LIDAR semantic segmentation\\ in autonomous driving.} &2022 \\
			
			\noalign{\smallskip}
			He et al.\cite{he2022robustDecision} & \makecell{Robust \\reinforcement \\ learning} & \makecell{SUMO} & \makecell {Decision making on \\highway entrance ramps} & \makecell{Proposing constrained adversarial \\strategies for self-driving\\high-speed entrance ramps.} &2022 \\
			
			\noalign{\smallskip}
			Pan et al.\cite{pan2019riskaverse} & \makecell{Robust \\reinforcement \\ learning} & \makecell{TORCS} & \makecell {Control strategies \\for dynamical models} & \makecell{Risk-adversarial learning approach, \\incorporating risk-averse and \\risk-seeking mechanisms acting on\\ games between agents.} &2019 \\
			
			\noalign{\smallskip}
			Yue et al.\cite{yue2019domainrandomization} & \makecell{Domain \\randomization} & \makecell{GTA} & \makecell {Semantic segmentation of \\autonomous driving scenarios} & \makecell{A new method of domain randomiza-\\tion and pyramid consistency is \\proposed to learn models with high\\ generalization ability.} &2019 \\
			
			\noalign{\smallskip}
			Kontes et al.\cite{kontes2020high} & \makecell{Domain \\randomization} & \makecell{CARLA} & \makecell {ADAS \\obstacle-avoidance} & \makecell{More complex road and high-speed \\traffic situations are considered, \\and the sim2real transformation is\\ accomplished by training several \\variants of the complex problem \\using domain randomization.} &2020 \\
			
			\noalign{\smallskip}
			Pouyanfar et al.\cite{pouyanfar2019roads} & \makecell{Domain \\randomization} & \makecell{KITTI} & \makecell {ADAS \\obstacle-avoidance} & \makecell{Static domain randomization uses real\\ data to solve the end-to-end collision-\\free depth drive problem.} &2019 \\
			
			\noalign{\smallskip}
			Kim et al.\cite{kim2017end} & \makecell{Transfer \\learning} & \makecell{Kinect} & \makecell {End-to-end autonomous driving \\with lane semantic segmentation} & \makecell{A continuous end-to-end transfer\\ learning approach which uses\\ two transfer learning steps.} &2017 \\
			
			\noalign{\smallskip}
			Akhauri et al.\cite{akhauri2020enhanced} & \makecell{Transfer \\learning} & \makecell{deepDrive} & \makecell {Autonomous driving\\ domain transfer} & \makecell{Adding the idea of robust RL to\\ transfer learning, learning both\\ parameters and strategies, and transferring\\ the correspondence between the two \\to the execution environment.} &2020 \\
			
			\hline
		\end{tabular}
	}
\end{table*}

\subsection{Curriculum learning}
\begin{figure}[tb!]
	\centering
	\includegraphics[width=3.5in]{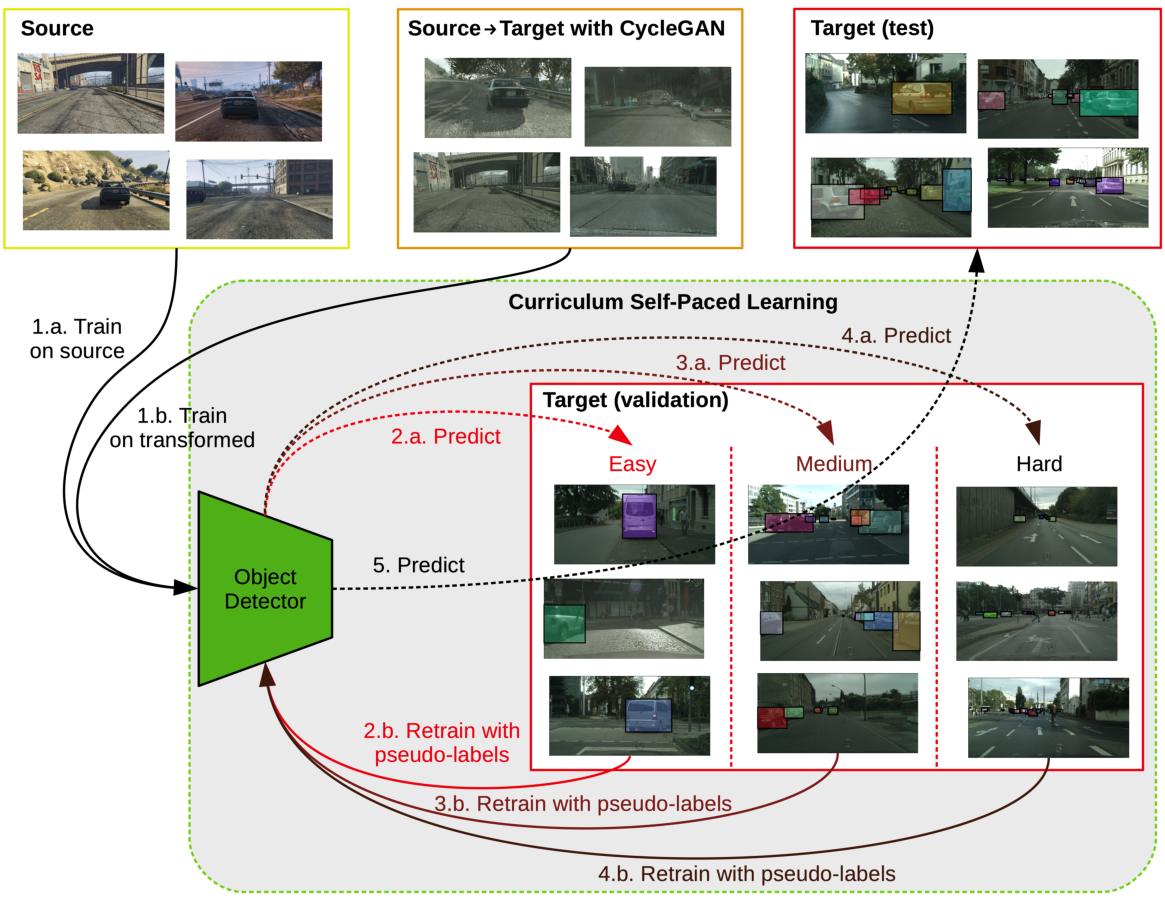}
	\caption{Curriculum self-paced learning approach for object detection in autonomous driving \cite{soviany2021curriculum}.}
	\label{fig_2}
\end{figure}

Curriculum learning (CL) \cite{bengio2009curriculumlearning} is a training strategy in which the model accumulates knowledge by initially learning simple tasks before involving more complex ones. It optimizes the sequence of accumulating the experience for agents to speed up the training process and improve effectiveness. The task of sequencing the samples is tedious, so Kumar et al. propose the concept of Self-Paced Learning \cite{kumar2010self}, where the curriculum is dynamically determined to adjust to the learning pace of the learner. A relevant method for object detection in autonomous driving tasks is proposed by Soviany et al.\cite{soviany2021curriculum}, which is shown in Fig. \ref{fig_2}. The model trains a target detector on the source images that are usually from the simulation environment, and then allows it to learn tasks from easy to hard levels by a self-paced learning method. Finally, it reaches the goal of predicting the images in the real domain in order to bridge the RG. However, adaptive methods need some prior knowledge and require manual design \cite{matiisen2019teacher}. In recent years, the method of combining of curriculum learning and reinforcement learning has been widely developed \cite{narvekar2020curriculum}. For example, Florensa et al. \cite{florensa2017reverse} propose a reinforcement learning-based curriculum learning method, which does not require priority knowledge and use inverse training, allowing the agent to directly start the curriculum from an applicable initial state. Nonetheless, this static approach sometimes does not work well in complex autonomous driving tasks.

To address complex driving tasks in dynamic scenarios, Qiao et al.\cite{qiao2018automatically} propose a dynamic idea named Automatically Generated Curriculum (AGC). They use deep reinforcement learning to develop a strategy and use candidate sets to generate the curriculum, which can optimize the traffic efficiency at complex intersections and reduce the training time. However, this approach of pre-training on other tasks and transferring the knowledge is not very effective in different driving situations. Bae et al.\cite{bae2021curriculum} propose a curriculum strategy with self-taught scoring functions, which effectively estimated the roll and sideslip angles of the vehicle in different driving situations. Meanwhile, Song et al.\cite{song2021autonomousovertaking} propose a high-speed autonomous overtaking system based on three-stages curriculum learning, using the same task-specific curriculum learning to train end-to-end neural networks, and experimental results proved that the method has better overtaking performance. Anzalone et al.\cite{anzalone2022end} propose a reinforcement learning method with five stages for end-to-end autonomous driving, following the idea of curriculum learning to continuously increase the learning difficulty of the agents at each stage to learn complex behaviors.

\subsection{Meta-learning}
The concept of meta-learning \cite{schmidhuber1987evolutionary}, which means learning to learn, has gained significant attention in recent years. The essence of meta-learning is that the model is expected to gain prior experience in other similar tasks and then be able to learn new knowledge more quickly. Meta-learning has been shown to have good advantages in several scenarios, such as single-task learning \cite{thrun1998learning}, multitask learning \cite{andrychowicz2016learning}, few-shot scenarios \cite{snell2017prototypical}, Neural Architecture Search (NAS) \cite{real2019regularized}, etc. Finn et al.\cite{finn2017model} propose a model-agnostic meta-learning algorithm which is compatible with all gradient descent, named as Model Agnostic Meta Learning (MAML). The model requires only a small amount of data to achieve fast convergence, and it addresses the previous drawback of focusing only on the initialization parameters in the moment.

%\begin{equation}
%\label{deqn_ex1a}
%\theta _{\rm{i}}^  = \theta  - \alpha \nabla _\theta  {\rm{L}}_{{\rm{T}}_{\rm{i}} } {\rm{f}}_\theta
%\end{equation}

\begin{figure}[tb!]
	\centering
	\includegraphics[width=3.5in]{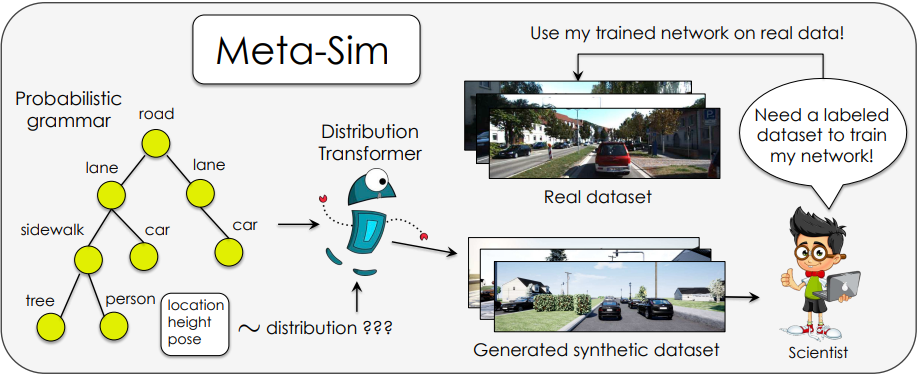}
	\caption{Synthetic datasets generation for driving scenarios based on the Meta-Sim method\cite{kar2019metasim}.}
	\label{fig_3}
\end{figure}

One of the challenges in traditional reinforcement learning is the inefficiency of data exploration. In order to solve the problem, Sæmundsson et al.\cite{saemundsson2018meta} refer meta-learning as a latent variable model and enable the knowledge to be transferred across the robotic system and automatically infer relationships between tasks from the data based on probabilistic ideas. In contrast, Nagabandi et al.\cite{nagabandi2018learning} propose an online adaptive learning approach for high-capacity dynamic models, where they implemented the algorithm on a vehicle to solve the sim2real problem by using model-based reinforcement learning with online adaptive methods while training a dynamic prior model using meta-learning. Jaafra et al.\cite{jaafra2019context} propose an adaptive meta-learning method based on an embedded adaptive meta reinforcement learning method with a neural network controller, which enables fast and efficient iterations for changing tasks and extends RL to urban autonomous driving tasks in the CARLA simulator. In addition, to reduce the high cost of the labeled datasets required to train a model with high performance, Kar et al.\cite{kar2019metasim} propose the concept of Meta-Sim, as shown in Fig. \ref{fig_3}, which is a learning model for generating synthetic driving scenarios with the goal of acquiring images through a graphics engine and their corresponding realistic images that bridge the distribution gap between the simulation and reality. Meta-Sim also optimizes a meta objective by automatically learning to synthesize labeled datasets related to downstream tasks.

\subsection{Knowledge distillation}
The concept of knowledge distillation was first introduced by Hinton et al.\cite{hinton2015distilling}. The main idea is that small student models usually learn from and are supervised by large teacher models for knowledge transfer. Knowledge distillation, as an effective technique for deep neural model compression, has been widely used in different areas of artificial intelligence \cite{krizhevsky2017imagenet}, including speech recognition, visual recognition, and natural language processing.

\begin{figure}[tb!]
	\centering
	\includegraphics[width=2.5in]{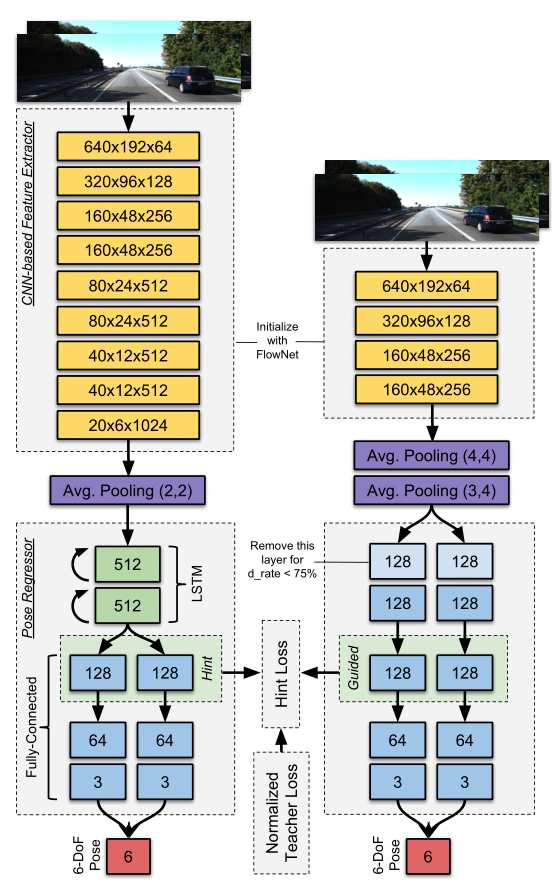}
	\caption{Network architecture of knowledge distillation for the student (right) and teacher (left) in \cite{saputra2019distilling}.}
	\label{fig_4}
\end{figure}

The knowledge transfer from the teacher to student is a critical component of knowledge distillation, but it is also a challenge to entirely learn from the real data for teacher models. To address this issue, Xu et al.\cite{xu2017training} introduce the conditional adversarial networks that enable the network to automatically learn to transfer knowledge from the teacher to student with smaller losses, thus make teacher-student models to learn better from real data. However, most of the teacher-student models are dependent on the teacher's predictions without considering ground truth error. In this case, Saputra et al.\cite{saputra2019distilling} suggest transferring knowledge only when the teacher is trusted, where the trust score is determined by the teacher's attrition. Fig. \ref{fig_4} shows the teacher-student structure with a distillation rate of 92.95$\%$ for trajectory prediction in autonomous driving. The teacher consists of a feature extractor network and a pose regressor network. By using the “dark knowledge” of the teacher, the student can transfer the knowledge from  simulation to reality. Similarly, Zhao et al.\cite{zhao2021sam} make the student in the model mimic the potential space of the teacher, and potentially represent the side information of the semantic segmentation in the teacher model, where the student tries to get a representation of the teacher input related to driving. Experimental results show that the model needs less labeling to achieve better results.

Object detection and semantic segmentation in point clouds obtained by LiDAR are important tasks for autonomous driving, and some researchers utilize the method of knowledge distillation to handle these tasks. Zhang et al.\cite{zhang2022pointdistiller} propose a knowledge refinement method based on point cloud graphs. The method extracts local geometric structures of the point cloud using dynamic graph convolution and weighted learning strategy to improve the efficiency of knowledge extraction, which is applied in sim2real for autonomous driving. Nevertheless, annotating some point cloud data requires significant cost. Sautier et al.\cite{sautier2022image} use a self-supervised knowledge distillation approach without any point cloud and image annotation, where the super pixels are used to extract features. To improve the robustness, Li et al.\cite{li2022selfDistillation} propose a robust LIDAR semantic segmentation method with Transformer-based Voxel Feature Encoder (TransVFE), where the encoder can learn features better by modeling and preserving local relationships between points.

\subsection{Robust reinforcement learning}
Robust reinforcement learning (RL) was proposed as a new paradigm for RL in \cite{morimoto2005robust} to represent the insensitivity of control systems to characteristic perturbations. The approach is defined based on $ H^\infty$ control theory and explicitly considers input disturbances and modeling errors to achieve robustness in RL against uncertainties that affect actions and states in the system, such as physical parameter changes.

\begin{figure}[tb!]
	\centering
	\includegraphics[width=3.5in]{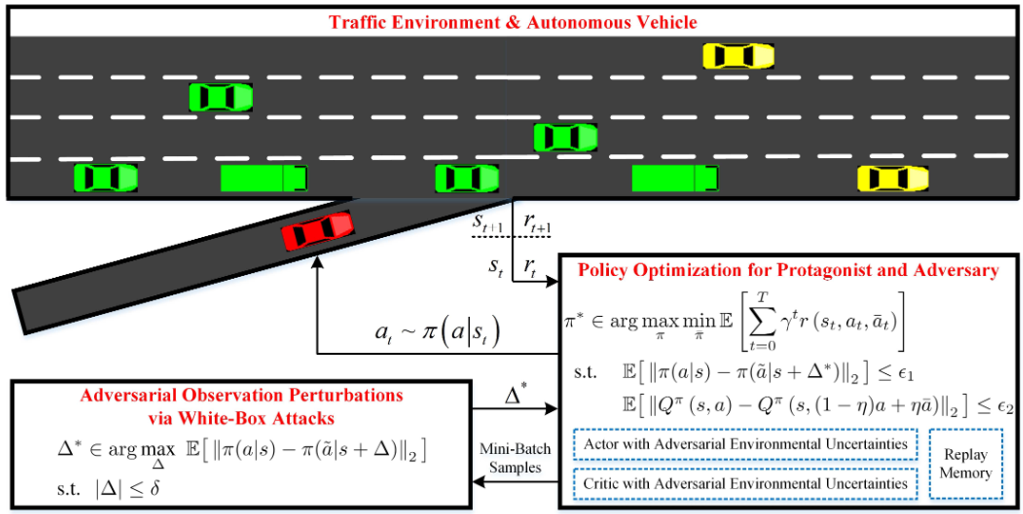}
	\caption{Decision making framework based on robust RL for highway on-ramp merging in autonomous driving \cite{he2022robustDecision}.}
	\label{fig_5}
\end{figure}

In the process of reinforcement learning research, classical techniques for improving robustness are shown to only prevent common situations and inconsistent for different simulation environments, so robust adversarial reinforcement learning (RARL) \cite{pinto2017robust} is proposed to improve the agents' behaviors by training the adversary to effectively prevent against system perturbations. As shown in Fig. \ref{fig_5}, a constrained policy for highway entrance ramps in autonomous driving is proposed by He et al.\cite{he2022robustDecision}. They model the environment of a highway intersection as an adversarial agent to constrain vehicle behaviors, and use adversarial agent and white-box adversarial attack techniques to simulate or generate adversarial environmental disturbances and observational perturbations so as to bridge the RG. In addition, authors also propose an observation-based adversarial reinforcement learning that influences the observations by incorporating a Bayesian optimization black-box attack method to make the agent efficiently approximate the optimal adversarial perturbation strategy \cite{he2022robustlane}. Pan et al.\cite{pan2019riskaverse} also propose a risk-adversarial learning approach, which incorporates a risk-averse mechanism by playing a game between risk-averse and risk-seeking agents while modeling risk as a value function. Experimental results show that this method triggers much fewer catastrophic events by agents than classical reinforcement learning methods.

Most of the existing literature refer RARL as a zero-sum simultaneous game with a Nash equilibrium, which may ignore the sequential manner of RL deployment, produce overly conservative agents, and lead to training instability. To handle this issue, Huang et al.\cite{huang2022robustreinforcement} present a new sequential robust RL formulation for both single-agent robot control and multi-agent highway merging in the task, where they propose a general and Stackelberg game model called RRL-Stack and developed the Stackelberg policy gradient algorithm to solve the RRL-Stack, which is better than the general RL method.

\subsection{Domain randomization}
Domain randomization is a method to solve the data augmentation problem for sim2real transfer. The idea is to train the data in the virtual domain, obtain the data distribution under various conditions in the virtual domain by sampling a set of random parameters, and then get the data results in the real domain and adjust the parameters to learn the sampling method that can be better generalized to the real domain in order to reduce the RG.

In the fields of autonomous driving and robotics, it is often necessary to deal with RGB images in a simulation environment. Tobin et al.\cite{tobin2017domainrandomization} directly transfer the strategy on simulated RGB images to the real world by domain randomization without real images for the first time. From the view of a stochastic algorithm, the method in \cite{sadeghi2016cad2rl} shows how the source domain can be properly randomized, by which authors train a generic model for different target domains. For some parameters of interest such as friction and mass, the method proposed in \cite{yu2017preparing} trains a generic control strategy through a dynamic model with calibrated parameters, and then analyzes the parameters using an online recognition system. Besides, Yue et al.\cite{yue2019domainrandomization} propose a new method of domain randomization and pyramid consistency to learn models with high generalization ability using consistency forced training of crossing-domain. The overall experiments show the effectiveness of domain randomization is much better than existing methods.

\begin{figure}[tb!]
	\centering
	\includegraphics[width=3.5in]{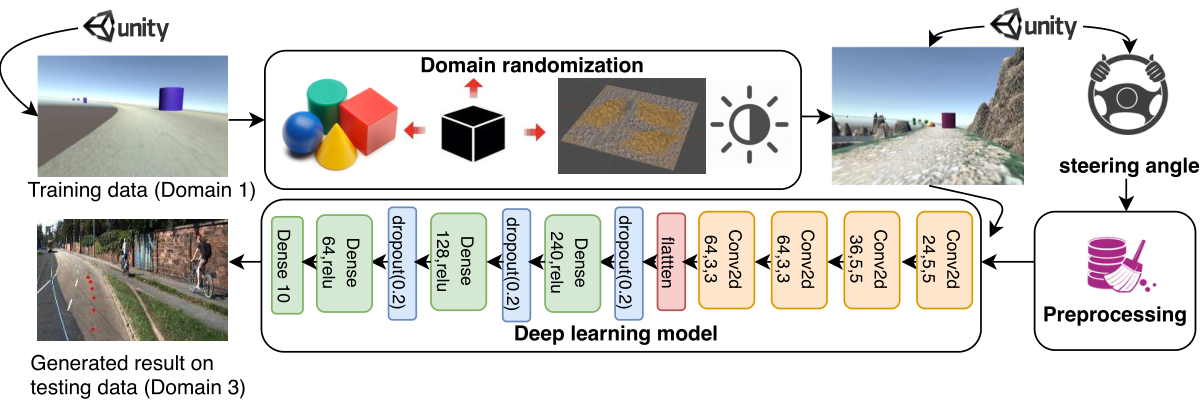}
	\caption{Framework for collision-free autonomous driving based on domain randomization \cite{pouyanfar2019roads}.}
	\label{fig_6}
\end{figure}

Kontes et al.\cite{kontes2020high} consider more complex road and high-speed driving situations by training several variants of the complex problem, in which authors use domain randomization to complete the sim2real transfer and transfer the learned driving strategies to real vehicles. In addition, to perform randomized simulation of visual features in the source domain, the method in \cite{mordatch2015ensemble} generates robust motion trajectories by randomized dynamics models and acts on real robots to guide them to perform learning multiple motor skills. Pouyanfar et al.\cite{pouyanfar2019roads} propose a dynamic domain randomization, which differs from the static domain randomization and contains the randomization of dynamic objects. The proposed framework based on domain randomization for collision-free autonomous driving is shown in Fig. \ref{fig_6}. It generates a simulated world by simulating vehicle driving to collect images and corresponding steering angles, and eventually predicts future steering angles in real images by scenario randomization, which solves the end-to-end collision-free driving problem using the real data.

\subsection{Transfer learning}
Transfer learning \cite{weiss2016survey} is a popular approach that has been developed to address the challenges associated with the high cost and time needed to label data in machine learning applications, especially supervised learning. The main idea of transfer learning is to transfer labeled data or knowledge between related domains in order to improve the learning effect of the target domain or tasks, in which the model can transfer from the existing labeled data (source domain data) to unlabeled data (target domain data). Transfer learning in autonomous driving is mainly used to implement the process of transfer knowledge learned from a source domain to another target domain, such as transferring driving strategies learned in a simulation environment to a real environment.

The high cost of information acquisition for autonomous driving has led to the consideration of collecting data in a simulation environment and transferring them to the real world in a reasonable manner. Isele et al.\cite{isele2017transferring} model intersection simulations and transfer their learned information to the real world, which demonstrates the robustness of the transfer process. Kim et al.\cite{kim2017end} propose a continuous end-to-end transfer learning approach that uses two transfer learning steps to transfer the environment and information to the nearby value domains of the vehicle to implement the learning process in a stepwise approach.

\begin{figure}[tb!]
	\centering
	\includegraphics[width=3.5in]{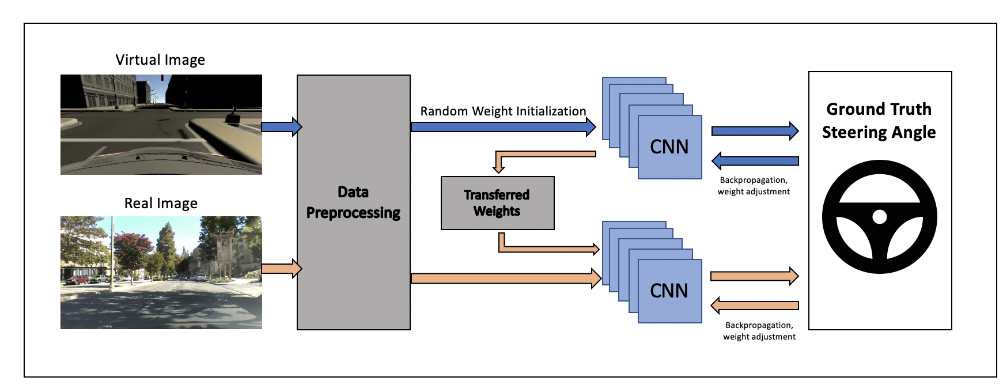}
	\caption{Transfer learning-based driving model including two stages with the transferred weights technique \cite{akhauri2020enhanced}.}
	\label{fig_7}
\end{figure}

To deal with the data acquisition problem, Tzeng et al.\cite{tzeng2020adapting} propose an improved adaptive approach that does not require manual labeling of the data and effectively compensates for the domain bias generated in transfer learning by weakly pairing the source-target approach. Gradient strategies have proven to be very effective in solving complex behavioral planning, and the multi-task policy gradient learning approach further improves the learning efficiency by using the transfer learning process among different tasks \cite{ammar2014online}. The same idea of robust RL can also be added to transfer learning by learning both parameters and policies in the process and transferring the corresponding relation between them to the execution environment \cite{akhauri2020enhanced}, whose network architecture is shown in Fig. \ref{fig_7}. The experiment conducted in the scene of high-speed intersections is divided into two stages. In the first stage, the simulated data are trained, and the transfer learning weight is used to train the real data in the second stage. To address the problem of the large network size that leads to poor program performance, Sharma et al.\cite{sharma2019semantic} use a pre-trained segmentation network, DeconvNet, where a variety of synthetic data and real data are used. The model achieves good results in simulating off-road driving environment.

\section{Digital twins technologies and applications}
While much progress has been made in sim2real in autonomous driving, transferring driving strategies from virtual simulation to the real world remains a challenge. It is time consuming and expensive to tune related algorithms to validate design requirements in real physical environment with various scenarios. They are trying to leverage more comprehensive simulations to reduce the number of places requiring physical tuning in industry.
\begin{figure}[tb!]
	\centering
	\includegraphics[width=3.5in]{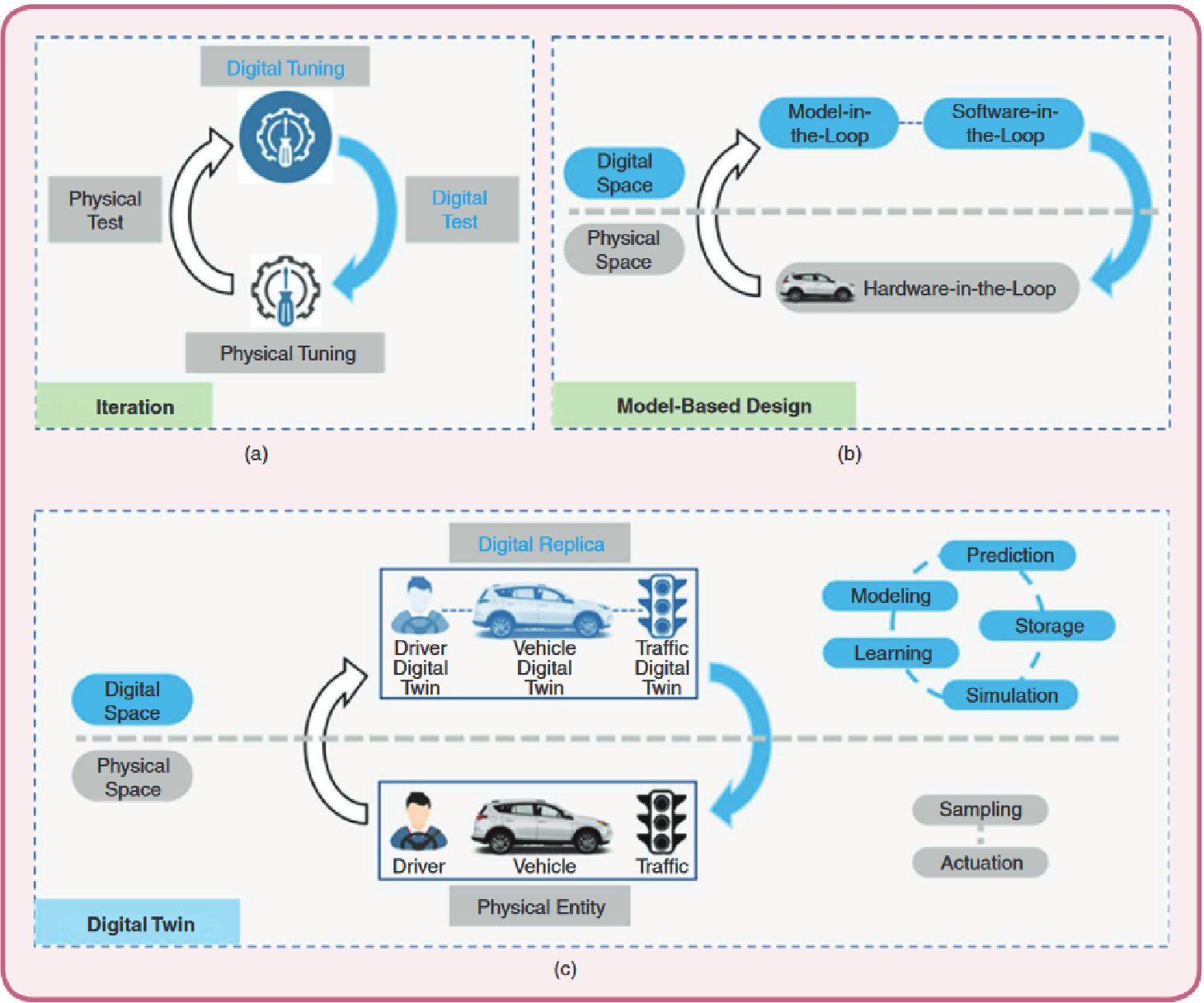}
	\caption{Comparison of digital twins with other simulation methods \cite{schwarz2022role}. (a) Iterative convergence, switching between physical and digital space, (b) Model-based design, digital and incremental exchanges, and (c) Digital twin, visual synchronization of data.}
	\label{fig_8}
\end{figure}

Virtual reality simulation has become a promising direction in various fields. The main idea of DTs, as an important embodied part of VR, is a virtual copy or model of any physical entity (physical twin), and both of them are interconnected by real-time data exchange \cite{singh2021digital, Vered2023The}. Conceptually, the digital twin technology is an extension of augmented reality and mixed reality. Meanwhile, parallel technologies provides the theoretical basis for the development of DTs.  Fig. \ref{fig_8} illustrates the differences between the digital twin method and its previous methods in autonomous driving, such as the iterative method (a typical method for creating virtual environment and tuning models) and the model-based design method \cite{schwarz2022role}. For a more comprehensive decomposition of DTs, Tao et al.\cite{tao2022digital} propose the five-dimensional digital twins model, which is presented as Eq. 1.

\begin{equation}
\label{deqn_ex1a}
M_{DT} = (PE,VM,Ss,DD,CN)
\end{equation}
where \emph{PE, VM, Ss, DD,} and \emph{CN} denote the physical entity, virtual model, service, data, and connection, respectively.

The AR, MR, PI and other technologies mentioned above have a profound impact on the simulation work and are inspirational for the applications of DTs in the autonomous driving field. Nowadays, there has been little work to summarize DTs and VR-related technologies in autonomous driving. Therefore, we review the methods and technologies from AR/MR, and PI to DTs and their related applications in autonomous driving in this paper.

\subsection{AR/MR-based methods and applications}
Augmented reality is a technology that enhances the perception of the physical world by integrating virtual objects into image sequences obtained from various cameras, allowing users to interact with these objects in real-time \cite{billinghurst2015survey}. The definition of AR has three characteristics including tracking, display, and interaction \cite{azuma1997survey}. The three characteristics also define the technical requirements of the AR system, namely that it needs a display that can display real and virtual images, a tracking system that can find the position of users' viewpoint and fix the virtual image in the real world, and a computer system that can interact with the user in real-time. Mixed reality, which includes both augmented reality and augmented virtual, introduces realistic scene information in the virtual environment through immersive ways, accompanied by interactive feedback networks to enhance environmental realism \cite{costanza2009mixed}. MR emphasizes the integration of real objects into the virtual environment, so the simulation ability in the framework is more crucial compared with AR. It creates a space where virtual and real elements coexist, and both can be allowed to interact with each other. As an off-line version of the digital twin system, MR can bridge the gap between the simulation and reality by enabling seamless interaction between physical and virtual objects that exist in a physical or virtual environment \cite{hoenig2015mixed}.

\begin{figure}[tb!]
	\centering
	\includegraphics[width=3.5in]{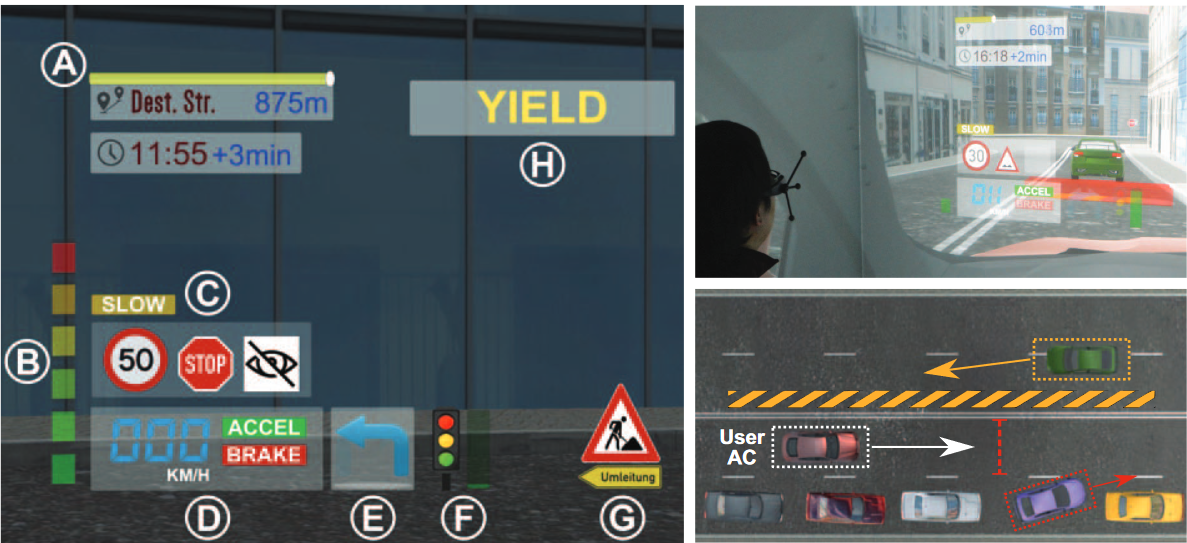}
	\caption{A windshield display interface prototype in a virtual environment for an emergency scenario \cite{lindemann2018explanatory}. Left: the display showing the viewport space and dynamic elements. Upper right: the prototype running in a virtual environment. Upper left: an example of requiring an emergency response.}
	\label{fig_9}
\end{figure}

In recent years, AR and MR have been used and developed in many fields such as assisted driving and autonomous driving. Lindemann et al.\cite{lindemann2018explanatory} design a windshield for AR effect display, which shows the perception and decision making of the autonomous driving system. As shown in Fig. \ref{fig_9}, the method provides the driver with information detected by the self-driving vehicle in the environment, such as other vehicles, obstacles, road information, and the driving decisions for assisting driving. The selected diversity and real-time interactive response ability of AR and MR in the virtual environment provide an excellent testing and validation platform for autonomous driving, which could be used to reduce the impact of RG. Su et al. \cite{su2018positioning} propose an optimized sensor array combined with the magnetic peg tracking technique in AR to track the position and direction of magnetic pegs in a magnetic field, which improves the positioning accuracy of an automated guided vehicle (AGV).  In \cite{szalai2020mixed}, the MR system uses sensors to capture vehicle state data and build a virtual space in order to provide a large number of optional complex environments for testing autonomous driving algorithms. Marc et al.\cite{zofka2018sleepwalker} propose a classical MR framework for autonomous driving based on LiDAR. The real sensors and sensor models both perceive their respective environments, and then the data perceived by each are fused into a mixed reality module to test the autonomous driving system that reacts to the real and virtual worlds.

The video perspective technologies can be extended to assist autonomous driving to enhance drivers' state perceptions, like highlighting pedestrians and vehicle density \cite{gao2022effects}. Through human-machine interface and internet of vehicles, Wang et al.\cite{wang2020augmented} propose an augmented reality-based assisted driving system for multi-vehicle interconnection. This method obtains the road, vehicle, and pedestrian information from multiple connected vehicles, collaboratively calculates and simultaneously displays this information through AR to assist drivers in driving. At the same time, AR makes it possible to visualize and interact with the data in DTs. Williams et al.\cite{williams2020augmented} use an AR-assisted DT solution, where digital data from the DT model are provided as a visual presentation through an AR device, and the DTs are used as a database to learn to generate motion commands for the vehicles in the virtual environment.

A typical approach of building MR systems with high-fidelity driving scenarios is to use high-precision simulators. Tettamanti et al.\cite{tettamanti2018vehicle} propose a vehicle traffic simulation system, which connects the SUMO simulator with real vehicles online to simulate various scenarios and provides data for vehicles in a virtual environment when testing autonomous driving methods. On top of that, Szalai et al.\cite{szalai2020mixed} propose a mixed reality virtual environment to verify the feasibility of building virtual environment through game engines such as SUMO and Unity3D, combined with a well-established system architecture. Mitchell et al.\cite{mitchell2019multi} propose an integrated system for MR with multiple intelligent agents, in which they add real vehicles to the environment and set up online learning strategies for the agents. The obstacle vehicles in the virtual environment are assigned simple tracing algorithms. This system provides a safe environment for testing autonomous driving models in the real world. Through the 3D location information provided by MR devices such as Hololens, Moezzi et al.\cite{moezzi2019autonomous} solve the simultaneous localization and map building problems and achieve path planning for intelligent vehicles.

MR can also be used as a system environment for DTs to digitally model unstudied objects and generate digital twin entities in a mixed reality environment. Tu et al.\cite{tu2021mixed} propose a method to enhance the mobility of the industrial crane digital twin through spatial registration and tracking in an MR system. To enhance the real-time performance and accuracy of the system distance measurement, Choi et al.\cite{choi2022integrated} propose an MR integration system combining deep learning and DTs, where the deep learning-based instance segmentation is applied in RGB and 3D point cloud data to improve registrations between the real robot and its digital twin. Lalik et al.\cite{lalik2021real} build a digital twin system integrated with existing control devices through human-robot interaction to improve the accuracy of robot motion, which is similar to the hardware-in-the-loop technology.

\subsection{Digital twins in autonomous driving}
The concept of digital twins has gained significant attention in Industry 4.0 due to its competitive and economic advantages. In the field of autonomous driving, digital twin technologies provide a safe and diverse testing environment with reducing its time cost by at least 80$\%$. At the same time, it solves two problems of traditional simulators \cite{yu2022autonomous}, \cite{salvato2021crossing}: 1) simulation testing is not equivalent to real end-to-end testing, and 2) factors such as weather, climate, and lighting are not fully covered. The digital twin paradigm is able to generate high-fidelity complex virtual environments and sensor modeling that approximates real data to realize a complete, comprehensive, accurate, and reliable digital twin entity.
%The framework of the digital-twins-based networked autonomous driving test environment is shown in Fig.10., which requires: i) the acquisition of real scene data, and ii) the acquisition of real vehicle data, both with different modeling methods and data processing.
Tab.2 provides the review of digital twins for autonomous driving applications.

\begin{table*}[!tbp]%调节图片位置，h：浮动；t：顶部；b:底部；p：当前位置
	
	\caption{AUTONOMOUS DRIVING APPLICATIONS STATISTICS FOR DIGITAL TWINS.}
	\label{tab:2}
	\resizebox{\linewidth}{!}{
		{
			\setlength{\tabcolsep}{8pt}
			\begin{tabular}{c c c c c}%表格中的数据居中，c的个数为表格的列数
				\hline\hline\noalign{\smallskip}	
				Article & Methodologies & Twin objects & Applications & Year \\
				\noalign{\smallskip}\hline\noalign{\smallskip}
				Laschinsky et al.\cite{laschinsky2010evaluation}  & MBD & Vehicles &ADAS design and test & 2010\\
				\\[1pt]
				Shikata et al.\cite{shikata2019digital}  &MBD & Vehicles &EV charging design and test& 2019\\
				\\[1pt]
				Szalai et al.\cite{szalai2020mixed}  & MBD, VPG & Vehicles, traffic &\makecell{ADAS test, \\mixed-reality application}& 2020\\
				\\[1pt]
				Dygalo et al.\cite{dygalo2020principles} & MBD & Vehicles & ADAS recognition & 2020\\
				\\[1pt]
				Wu et al.\cite{wu2021digital}  &MBD &Environment&\makecell{Model-based RL \\in autonomous driving}
				& 2021\\
				\\[1pt]
				Yu et al.\cite{yu2022autonomous}  &MBD, CPS & Vehicles, environment &ADAS design and test, V2X& 2022\\
				\\[1pt]
				Schwarz et al.\cite{schwarz2010digital}  &CPS & Vehicles, environment &ADAS design and test & 2010\\
				\\[1pt]
				Eleonora et al.\cite{bottani2017cyber}  &CPS & Vehicles &AGV logistics action test& 2017\\
				\\[1pt]
				Chen et al.\cite{chen2018digital}  &CPS & Drivers & Drivers' safety behaviors analysis& 2018\\
				\\[1pt]
				Rassõlkin et al.\cite{rassolkin2019digital}  &CPS, MBD & Vehicles &ADAS test& 2019\\
				\\[1pt]
				Ge et al.\cite{ge2019research}  &CPS & Vehicles &ADAS Design and test, V2X& 2019\\
				\\[1pt]
				Veledar et al.\cite{veledar2019digital}  &CPS & Vehicles &ADAS design and test & 2019\\
				\\[1pt]
				Liu et al.\cite{liu2020sensor} & CPS & Vehicles & \makecell{Multi-sensor fusion for \\vehicle recognition, ADAS} & 2020\\
				\\[1pt]
				Liu et al.\cite{liu2021towards} & CPS & Vehicles & ADAS design and test & 2021\\
				\\[1pt]
				Culley et al.\cite{culley2020system} & VPG, MBD & Vehicles, environment & ADAS design and test & 2020\\
				\\[1pt]
				Fremont et al.\cite{fremont2020formal}  & VPG & Environment, vehicles &Formal test based on scenario& 2020\\
				\\[1pt]
				Voogd et al.\cite{voogd2022reinforcement}  &Transfer learning &Vehicles&\makecell{Reinforcement learning\\ autonomous driving}& 2022\\
				\\[1pt]		
				\noalign{\smallskip}\hline\noalign{\smallskip}
				
				\multicolumn{5}{c}{MBD: model-based design, CPS: cyber-physical systems, VPG: virtual proving ground.} \\
				\noalign{\smallskip}
				\hline
			\end{tabular}
		}
	}
\end{table*}

%\begin{figure}[!h]
%	\centering
%	\includegraphics[width=3.5in]{fig9}
%	\caption{Automatic driving test framework based on digital twins.}
%	\label{fig_10}
%\end{figure}

\begin{figure}[tb!]
	\centering
	\includegraphics[width=3in]{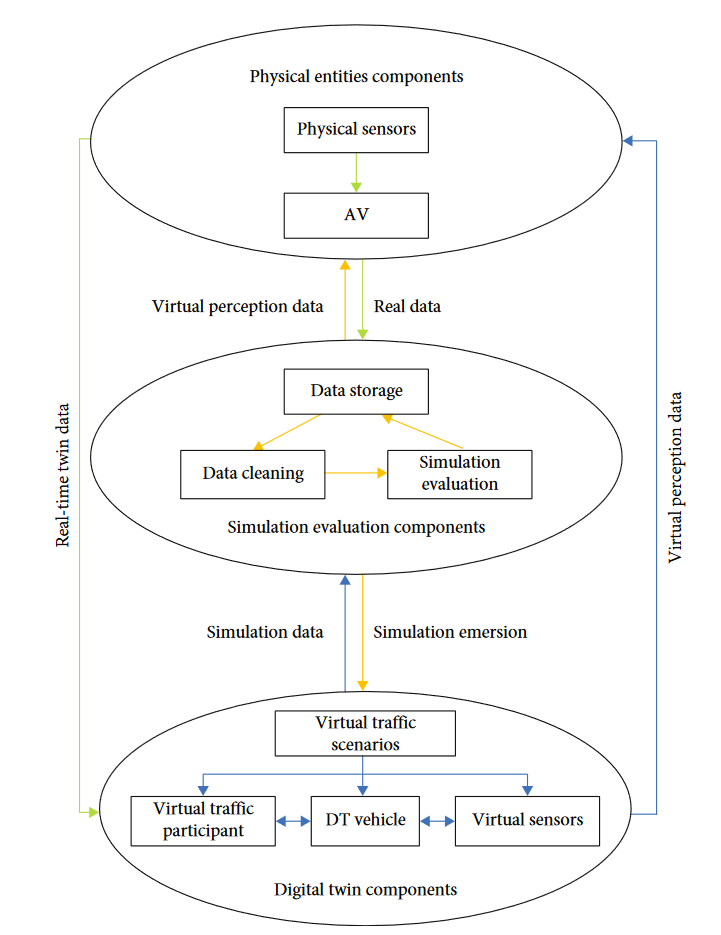}
	\caption{Block diagram of data flow based on the digital twin model. The green, blue, and yellow lines indicate the physical entity data stream, the DT stream, and the simulation evaluation data stream, respectively \cite{xiong2022design}.}
	\label{fig_11}
\end{figure}

In the field of autonomous driving, a digital twin system is usually constructed with three parts \cite{yu2022autonomous}: 1) the digital twin of the sensor model, 2) the digital twin of 3D digital maps, and 3) the logical twin corresponding to traffic flow control. Researchers typically design task-specific digital twin systems based on the above points. The digital twin system for automated guided vehicles enables the autonomous transportation of driven robots among three predefined tracks of processing stations through digital twin modeling, where the digital twin is only used as a dynamic simulation of discrete events without involving complex dynamic driving control techniques \cite{bottani2017cyber}. Rassõlkin et al.\cite{rassolkin2019digital} propose a digital twin system for autonomous electric vehicles. They use matlab to model the sensors and vehicle models of the ISEAUTO vehicle and then combine data from the real devices and virtual sensors with a machine learning program.

In recent years, many different frameworks have been proposed for the digital twin in autonomous driving. These digital twin frameworks are generally divided into physical, data, and virtual parts, and the users can add additional modules in order to accomplish different tasks. Xiong et al. \cite{xiong2022design} present a physical entity simulation framework, as shown in Fig. \ref{fig_11}, which consists of three parts: the physical entity, the digital twin component, and the simulation evaluation component. The physical entity part is responsible for receiving real-world data and executing relevant operations. The simulation evaluation component is responsible for processing the received real and virtual data, and the digital twin component performs the simulation verification of the DT vehicles. The Society for Automotive Engineering (SEA) also proposes a framework consisting of four modules: data, integration, analysis, and digital twin. Each module corresponds to a different task and meets the requirements of digital twin components, which directly interact with the physical system to achieve autonomous driving tasks \cite{almeaibed2021digital}.

Vehicle to everything (V2X) based on vehicle networking is also playing an important role in the study of the digital twins for autonomous driving. China Academy of Information and Communication (CAICT) develops a digital twin system that uses the V2X communication technologies to achieve high-fidelity transmission and simulation, which is used in an autonomous driving testing system \cite{ge2019research}. Niaz et al.\cite{niaz2021autonomous} then confirm that the V2X-based DT system supports low latency connectivity for autonomous driving tests. Wang and Tu \cite{wang2022automatic} propose a method that can automate traffic modeling to automatically capture and reconstruct the traffic in special areas. This method greatly enhances the traffic flow of autonomous driving and improves the effectiveness of the algorithm. A framework for the application of vehicle edge computing in the DT is proposed to model the physical vehicle environment through edge heterogeneous information obtained by vehicles in the DT, combining with communication techniques of vehicle-to-infrastructure (V2I) \cite{xu2022enabling}.

The high-fidelity simulator LGSVL \cite{rong2020lgsvl} for autonomous driving provides an end-to-end, full-stack simulation, through which users can create required digital twin entities, including customizable vehicle sensors and core modules, to achieve the requirements of autonomous driving algorithms. Liu et al.\cite{liu2020sensor} propose a digital twin system based on the Unity game engine, which depicts the road traffic situation by integrating physical target detector data and digital twin information. They demonstrate and evaluate the approach in the Unity engine, and the results show that their method combined with the digital twin system can significantly improve driving safety.

In reinforcement learning, DT technologies are used to improve the ability of transferring learned strategies and bridge the gap between different domains by the systematic and simultaneous combination of data from traffic scenarios in both virtual and real worlds. Voogd et al.\cite{voogd2022reinforcement} apply domain randomization and adaptive techniques at each stage of the development cycle of digital twins to make the transfer learning process more robust. The multi-modal fusion of driver digital twins (DDTs) is proposed in \cite{hu2022review}, which emphasizes the ability of the driver's intrinsic personality and extrinsic physiological level to bridge the gap between the real autonomous driving and fully digital systems.

\section{Parallel intelligence technologies}
Sim2real-based methods transfer the learned knowledge from the simulated to real worlds, and DT-based methods allow autonomous driving vehicles to learn knowledge by synchronizing data from both the real and simulated worlds in digital twins. Both of them can handle the RG problem to some extent, but they still face some challenges such as the difficulty in modeling complex systems. The RG is very large in some complex scenarios, to which the dynamic control simulation in DT and sim2real methods cannot adapt. In this case, it is an effective way to solve the RG problem to construct an artificial system, realize the dynamic interaction of computation, physics and society, and deal with uncertainty through parallel computation of artificial and real systems.

\subsection{Parallel intelligence concept and ACP method}
The parallel intelligence concept, which was first proposed by Wang \cite{wang2004parallel}, provides an effective idea to connect the physic world and artificial world and achieves a great success in many fields in recent years. This concept is based on the cyber-physical-social systems (CPSS) that is the evolution of cyber-physical-systems (CPS). CPSS enables virtual and real systems to interact, feedback, and promote each other. The real system provides valuable datasets for the construction and calibration of the artificial system, while the artificial system guides and supports the operation of the real system, thus achieving self-evolution.  Alam and Saddik \cite{alam2017c2ps} propose a cloud-based cyber-physical system (C2PS) intelligent interaction controller, which incorporates cloud technologies in a cyber-physical system. The architecture uses Bayesian networks to systematically and dynamically consider the context for prototype telematics-based driver assistance applications in the vehicle Cyber-Physical Systems (VCPS). Bhatti et al. \cite{bhatti2021towards} propose a framework for an IoT-based digital twin for electric vehicles, where they explain the detailed roles of IoT and the DT technologies in autonomous navigation control, advanced driver assistance systems and vehicle health detection for electric vehicles.

To implement the parallel intelligence concept, the ACP method, an organic combination of artificial societies (A), computational experiments (C) and parallel execution (P), is proposed by Wang. et al. \cite{wang2008toward}, which has made great achievement in many fields. The ACP method consists of three main steps: 1) modeling complex systems using artificial systems, 2) using computational experiments to train and evaluate complex systems, and 3) setting the interaction between the real physical system and the virtual artificial system so as to realize effective parallel control and management of the complex system. Unlike  DT, which is still in the partial computational experiment stage of the ACP method and only establishes a descriptive artificial system without prediction and guidance functions, providing real-time monitoring and adjustment services for specific systems, PI provides a complete modeling and description of the ACP method with three functions of description intelligence, prediction intelligence, and prescriptive intelligence, so it can achieve better management and control of complex systems.

In recent years, researchers applied the ACP and parallel intelligence methods in handling knowledge transferring from the simulation to reality worlds. The parallel intelligence concept has also evolved in several technologies, such as parallel learning \cite{chen2018parallelmotion}, parallel vision \cite{wang2017parallelvision}, parallel control \cite{wang2010parallel, lu2022event}, parallel testing \cite{li2019parallel},  parallel driving \cite{liu2019research, yang2022parallel}, and parallel planning \cite{chen2018parallelmotion}, etc.  The survey of these methods is shown in Table \ref{tab:4}.

\begin{table*}[!tbp]%调节图片位置，h：浮动；t：顶部；b:底部；p：当前位置
	
	\caption{COMPARISON OF SELECTED VIRTUAL-TO-REAL WORKS ON CONTROL OF AUTONOMOUS DRIVING WITH REAL DATA.}
	\label{tab:4}
	\resizebox{\linewidth}{!}{
		{
			\setlength{\tabcolsep}{8pt}
			\begin{tabular}{c c c c c}%表格中的数据居中，c的个数为表格的列数
				\hline\hline\noalign{\smallskip}	
				Article & Virtual & Methodologies& Highlights & Year \\
				\noalign{\smallskip}\hline\noalign{\smallskip}
				Li et al.\cite{li2017parallel}  &Software Defined Artificial Systems & parallel learing&\makecell{A parallel learning framework for\\ extracting data and integrating learning}& 2017\\
				\\[1pt]
				Zhang et al.\cite{zhang2020parallelvision} & Pedestrians datasets&parallel vision&\makecell{Parallel vision for long-term\\ learning in an ever-changing environment}& 2020\\
				\\[1pt]
				Zheng et al.\cite{zheng2020novel}  & Changedetection dataset &parallel vision&\makecell{Background subtraction algorithm \\based on parallel vision and GANs}& 2020\\
				\\[1pt]
				Wang et al.\cite{wang2017paralleldrivinginCPSS}  &iHorizon &parallel driving &\makecell{Cooperative control of \\multi-vehicle autonomous driving}& 2017\\
				\\[1pt]

				Chen et al.\cite{chen2019learning}  &PED(end-to-end driving data set)& parallel driving&\makecell{Hybrid virtual-reality data for autonomous\\ driving performance improvement}& 2019\\
				\\[1pt]
				Liu et al.\cite{liu2020digital}  &PanoSim & parallel driving&\makecell{Parallel driving with safety\\ and smart connectivity}& 2020\\
				\\[1pt]

				Chen et al.\cite{chen2022parallel}  &PDOS(ubiquitous operating system)& parallel driving&\makecell{Parallel drive system \\for autonomous driving tasks}& 2022\\
				\\[1pt]
				Li et al.\cite{li2019parallel}  & Customizing environment &parallel testing&\makecell{Closed-loop parallel test system\\ for more challenging tests}& 2019\\
				\\[1pt]
				Chen et al.\cite{chen2018parallelmotion}  & GTA5/ETS2 &parallel planning&\makecell{End-to-end parallel planning \\ for emergency scenarios}& 2019\\
				\\[1pt]
				Liu et al.\cite{liu2022study}  &Customizing the UAV environment & parallel planning&\makecell{Parallel planning system\\ for UAV clusters}& 2022\\
				\\[1pt]
				Wang et al.\cite{wang2010parallel}  & ATS &parallel control&\makecell{Parallel traffic \\management framework}& 2010\\
				\\[1pt]
				\noalign{\smallskip}\hline\noalign{\smallskip}
			\end{tabular}
		}
	}
\end{table*}

\begin{figure}[tb!]
	\centering
	\includegraphics[width=3in]{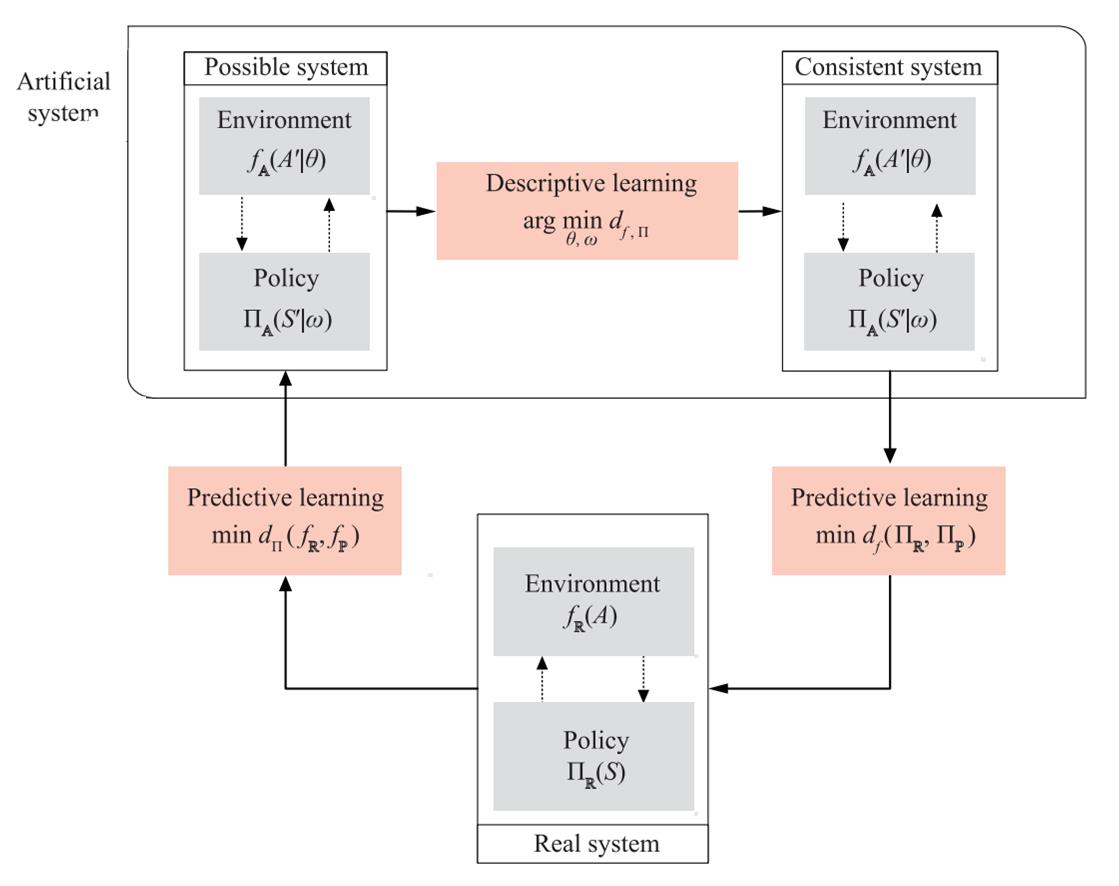}
	\caption{ Parallel learning framework \cite{li2017parallel}.}
	\label{fig_20}
\end{figure}

\begin{figure}[b]
	\centering
	\includegraphics[width=3in]{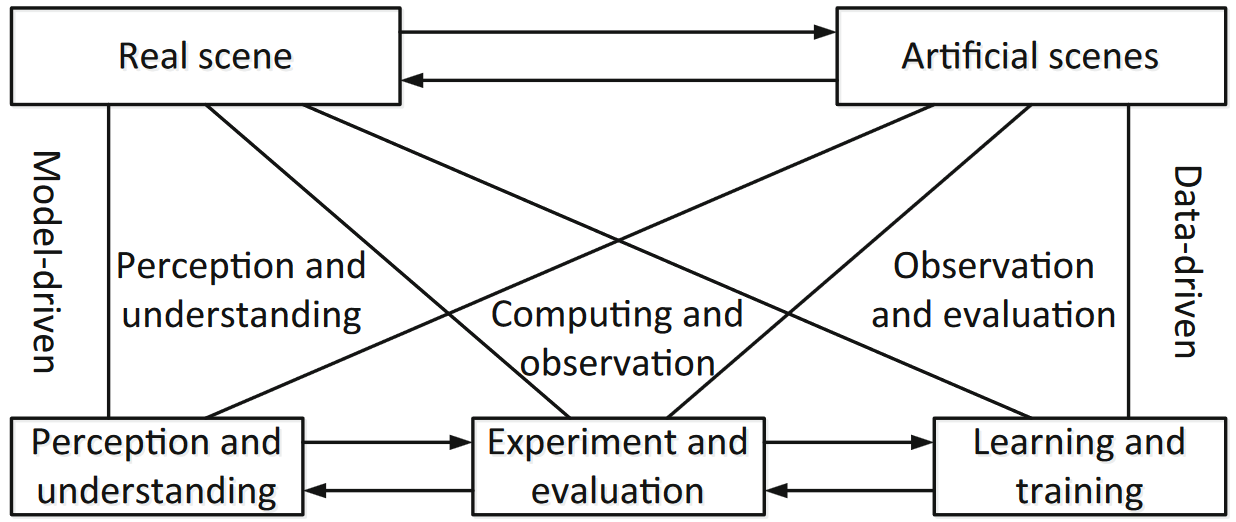}
	\caption{ Parallel vision framework \cite{li2017parallel}.}
	\label{fig_21}
\end{figure}

\subsection{Parallel learning and parallel vision}
After proposing the concept of parallel intelligence, researchers attempt to extend parallel intelligence into the machine learning field and construct a novel theoretical framework called parallel learning \cite{li2017parallel}. The framework of parallel learning is illustrated in Fig. \ref{fig_20}, which can be divided into three interrelated stages: descriptive learning, predictive learning, and prescriptive learning. Descriptive learning involves the extraction of knowledge and experience from extensive artificial data within an artificial system. Predictive learning aims to establish mappings from observed information to unknown information. Prescriptive learning further optimizes the learning strategy by leveraging these two processes while providing guidance to enhance the system's performance. Specifically, this process addresses the issue of inadequate recorded data resulting from manual operations by employing predictive learning with a substantial number of generated artificial data. Descriptive learning is subsequently employed to enhance the acquired knowledge and experience. The framework provides a new approach of handling the problems of reality gap and data imbalance from the real and simulated scenarios for machine learning-based autonomous driving methods.

Computer vision is one of the most significant technologies in machine learning and autonomous driving. Based on the ACP theory, Wang et al. \cite{wang2016parallel} propose the concept of parallel vision to address the challenge of extracting features from complex environments in traditional computer vision methods. As shown in Fig. \ref{fig_21}, the approach entails constructing artificial scenes that resemble real driving scenarios and automatically obtaining accurate labeling information while generating large-scale and diverse datasets. Subsequently, computational experiments are conducted in these artificial scenes, including two stages: ``learning and training" and ``experimentation and evaluation". In the learning and training stage, the model is trained using the generated large-scale and diverse datasets from both artificial and real worlds. In the experimentation and evaluation stage, the model performance is assessed in the two environments using the datasets. Computer vision models run in parallel in real and artificial driving scenarios, constantly fusing both of the two scenarios to improve the performance of knowledge transfer in complex environment to better reduce RG. On this basis, parallel vision paves the way for vision technologies in autonomous driving.

\subsection{Parallel driving and parallel testing}
\begin{figure}[tb!]
	\centering
	\includegraphics[width=3in]{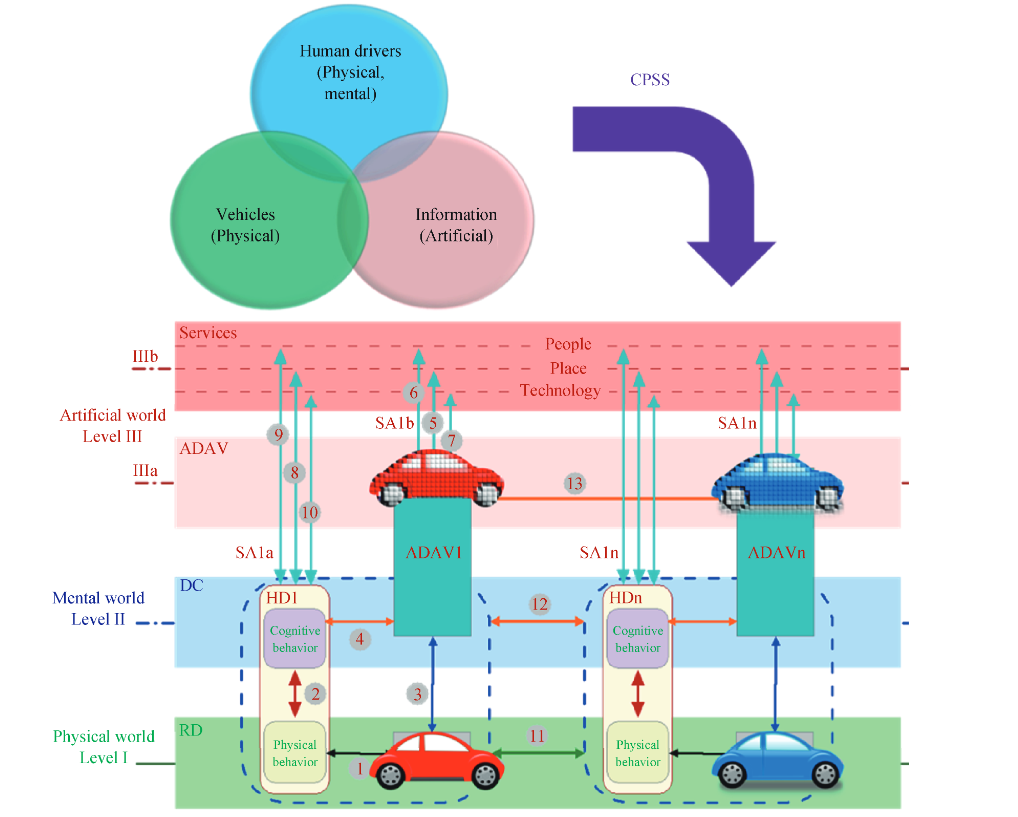}
	\caption{Framework of parallel driving. (RD: real driving; DC: driver cognition; CPSS services including three components: people (social web), place/location (geo web), and technology (sensors, Internet of Things, etc); HD: human driver; RV: real vehicle; ADAV: artificial driver and artificial vehicle; and SA: situation awareness) \cite{wang2017paralleldrivinginCPSS}.}
	\label{fig_17}
\end{figure}

The development of autonomous driving poses significant challenges to current vehicle and transportation systems. Wang et al. \cite{wang2017paralleldrivinginCPSS} propose three elements required for future connected autonomous driving: physical vehicles, human drivers, and cognitive attributes. Based on the ACP theory, the authors develop projecting the three elements to three parallel worlds and creating a parallel driving framework. As shown in Fig. \ref{fig_17}, the three parallel worlds are interconnected. Each vehicle in the real world is assigned controlling functions as well as an ADAV module which is responsible for communication with the artificial world and other vehicles. This module also provides driving information for human drivers. Facilitated by the ADAV module, parallel driving enables joint responses to complex autonomous driving scenarios, with human drivers conditionally participating to ensure system safety. Furthermore, Liu et al. \cite{liu2019research} integrate ``digital quadruplets" including the physical vehicle, the descriptive vehicle, the predictive vehicle, and the prescriptive vehicle in parallel driving. Based on the description of digital quadruplets, three virtual vehicles, which are defined as three ``guardian angels" for the physical vehicle, play different roles to make intelligent vehicles safer and more reliable in complex scenarios. Autonomous driving methods based on parallel driving can learn drivers' behavior in response to different reality and simulation scenarios through the integrated multi-ADAV modules so as to bridge the RG.

\begin{figure}[b]
	\centering
	\includegraphics[width=3in]{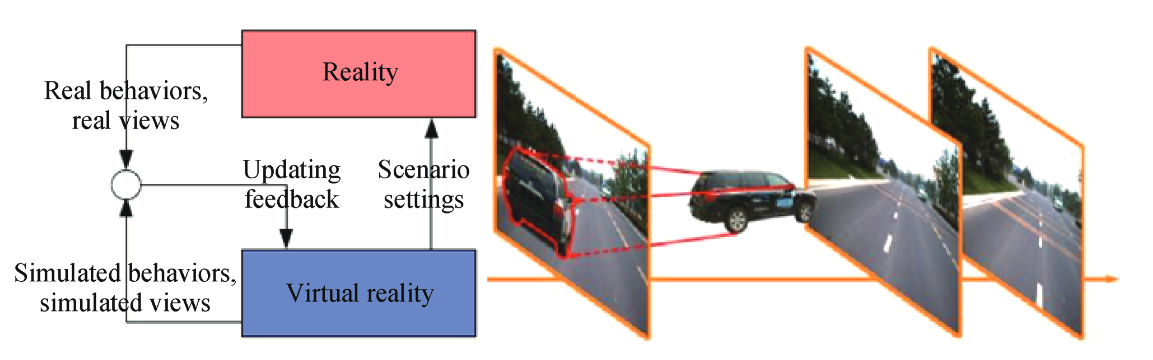}
	\caption{Realization of parallel testing \cite{wang2017paralleldrivinginCPSS}.}
	\label{fig_18}
\end{figure}

 %Additionally, the parallel testing system is a comprehensive system that contains parallel vision and parallel planning. It represents another important application of parallel systems in autonomous driving.

In the research of parallel driving, Wang et al. \cite{wang2017paralleldrivinginCPSS} propose the initial concept of parallel testing, in which the cyclic updating method is used to address the RG problem and verify the performance of autonomous driving. As shown in Fig. \ref{fig_18}, the cyclic updating method of co-evolution between the real testing ground and parallel virtual testing ground enhances the interaction of virtual reality and authenticity of the scenarios, so the test in virtual testing ground can show the performance of autonomous driving. Following this research, Li et al. \cite{li2019parallel} complete the framework of parallel testing which emphasized the real-time data interaction between real and virtual scenarios and the diversity of testing scenarios.

\subsection{Parallel planning and parallel control}
\begin{figure}[tb!]
	\centering
	\includegraphics[width=3in]{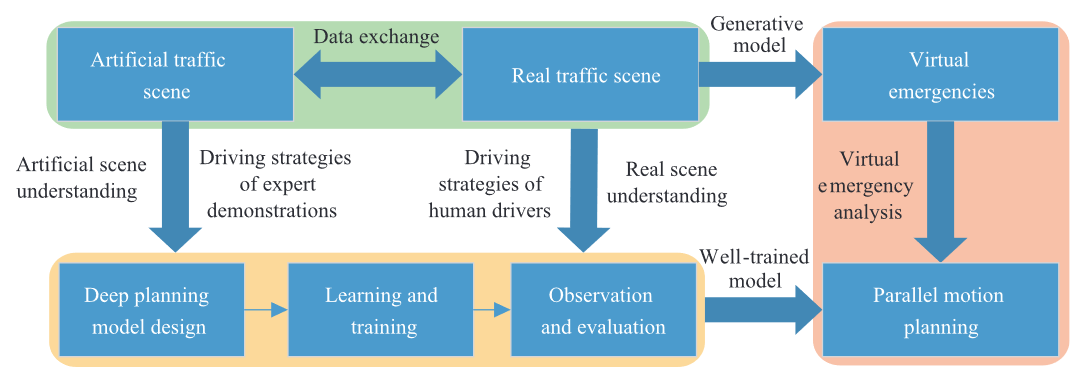}
	\caption{Framework of parallel motion planning. \cite{chen2018parallelplanning}.}
	\label{fig_16}
\end{figure}

Planning is one of the important parts for autonomous driving systems. To address the problem of emergency traffic scenarios in the real world for self-driving vehicles, Chen et al. \cite{chen2018parallelplanning} propose the method of parallel planning. As shown in Fig. \ref{fig_16}, the method involves modeling emergency traffic scenarios from the artificial traffic world, in which emergency traffic scenarios are generated for learning autonomous driving algorithms. The gained knowledge can enable correctly and timely planning in real-world scenarios when emergent events occur. Parallel planning integrates parallel perception, deep learning, and end-to-end autonomous driving into a framework, and incorporates knowledge guidance of emergency avoidance from expert strategies in the planning process, which combines with synthesized data from both virtual and real worlds. Based on this mode, it can effectively apply the knowledge learned in the artificial world, especially for the driving strategies for emergency scenarios, in the real world and enhance the applicability of artificial driving behaviors to real-world situations.

Parallel control is a knowledge-based automated learning way that aims to achieve automation of intelligent systems and simplification of complex systems. Wang et al.\cite{wang2006modeling} first propose parallel control framework based on the ACP theory. %, as shown in Fig. \ref{fig_15}. This method first constructs an artificial system based on collected data, and then generates a large amount of artificial data to optimize control in the artificial system. Finally, the control strategy of the artificial system is used to guide the control and optimization of the real control system. The whole process involves the interconnection between the real and artificial systems, comparing their control strategies to evaluate future states. Feedback is then provided to adjust their respective management and control methods and optimize the control of complex real systems.
In recent years, parallel control is used in transportation systems to construct a parallel transportation management system and verify its effectiveness in complex transportation systems \cite{wang2010parallel}.  Based on the synthetical data and real-world observations, parallel control enhances control effectiveness and improves the development of autonomous driving systems.

\section{Autonomous driving simulators}
It requires significant costs to directly test autonomous driving algorithms in real world scenarios, and the data of rare events in the real world are difficult to collect. Some researchers devote themselves to construct parallel datasets \cite{8913579}, but the collected off-line datasets cannot meet the requirements of real-time interactive testing. In this case, low-cost testing of autonomous driving systems, simulation systems are often built to simulate testing processes. Autonomous driving simulators need to be able to build realistic road environment as well as vehicle models, sensors, such as cameras, and LiDAR.

\begin{table*}[!tbp]%调节图片位置，h：浮动；t：顶部；b:底部；p：当前位置
	\centering
	\caption{SIMULATOR AND RELATED DESCRIPTIONS FOR AUTONOMOUS DRIVING.}
	\label{tab:3}
	\resizebox{\linewidth}{!}{
		\begin{tabular}{c  c  c }%表格中的数据居中，c的个数为表格的列数
			\hline\hline\noalign{\smallskip}	
			Platform & Descriptions & Advantages\\
			\noalign{\smallskip}\hline\noalign{\smallskip}
			\\[0.5pt]
			AirSim \cite{shah2018airsim} & High fidelity platform developed by unreal engine & High-fidelity environment and car models\\
			\\[0.5pt]
			Gazebo \cite{koenig2004design} & Open source robot simulator & Strong physical modeling capabilities\\
			\\[1pt]
			CARLA \cite{dosovitskiy2017carla} & Unreal engine open source autopilot simulator & High-fidelity sensor data\\
			\\[1pt]
			LGSVL \cite{rong2020lgsvl} & Autonomous robot car simulator for unity game engine & \makecell{High-precision maps and interfaces to support\\ multiple autonomous driving algorithm engines}\\
			\\[1pt]
			Torcs \cite{wymann2000torcs} & Vehicle driving games & High portability and modifiability\\
			\\[1pt]
			MetaDrive \cite{li2022metadrive} & Open source driving simulation platform & \makecell{Lightweight, allowing intensive learning \\training using a large number of random maps}\\
			\\[1pt]
			SUMO \cite{krajzewicz2010traffic} & Open source traffic simulation platform & Generate complex traffic systems\\
			\\[1pt]
			SUMMIT \cite{cai2020summit} & CARLA-based traffic environment extension & \makecell{Generate complex traffic systems \\and can import the SUMO mapping system}\\
			\\[1pt]
			AutoWare \cite{kato2018autoware} &Open-source autonomous driving platform & \makecell{Mature development framework}\\
			\\[1pt]
			Apollo \cite{huang2018apolloscape} & Large-scale datasets consisting of video and 3d point clouds & \makecell{Data for high precision attitude information}\\
			\\[1pt]
			\noalign{\smallskip}\hline
		\end{tabular}
	}
\end{table*}

Simulators are usually developed based on game engines with certain rendering capabilities, and most are static simulators such as Unreal, and Unity, which have limited control over characters and weather. However, simulators that focus on the task of driving vehicles usually need to build dynamic scenes. For the standard of scene simulation, the open series developed by the Association for Standardization of Automation and Measuring Systems(ASAM) is developing as a guide for autonomous driving scenarios. A typical software based on OpenScenario is RoadRunner Scenario, which is an interactive dynamic scenario simulator. Besides, CARLA\cite{dosovitskiy2017carla} and LGSVL \cite{rong2020lgsvl}, etc. can also be used to build driving scenarios based on OpenScenario for autonomous driving tasks. In addition, some simulation environments improved based on robot simulators like Gazebo \cite{koenig2004design} can achieve more details at the unit level such as vehicle models, pedestrian modeling, and higher accuracy for sensors, but not realistic enough for 3D scene implementation. Currently, some high-fidelity simulators have been developed as virtual environment for digital twins in autonomous driving, such as AirSim \cite{shah2018airsim}, CarSim \cite{benekohal1988carsim}, and SUMO \cite{krajzewicz2012recent}. Tab.3 provides the review of the simulators.

\section{Problems and challenges}
Although the methods of sim2real, DTs and PI have made great achievements in the field of autonomous driving, it still needs to further improve existing limitations and explore unknown technologies in the process of developing a reliable, safe and comfortable autonomous driving method in the real world. It is necessary for researches to focus the work on sim2real, DT, and PI technologies and consider how to bridge the RG, in order to promote the development of extensible and safe autonomous driving. At present, there are four main challenges in this field.

(1) Current methods and technologies are designed for specific scenarios, but their performance applied in other different tasks is not satisfactory. There is a lack of a general task-independent method to ensure the effectiveness and portability of the sim2real process. In this case, we need to consider a way to quantify and represent the reality gap in the current autonomous driving field and compare different environments and applications, which is conducive to further research. Moreover, quantifying RG is deriving relevant algorithms to reduce RG and achieve the purpose of leaping from the virtual to real world.

(2) As a solution for training generalized reinforcement learning models, sim2real needs a lot of data to support its training process. We expect to use more real data, but real data are difficult to obtain compared with simulated data. Real data are often difficult and challenging to collect, classify, and generalize to a standard distribution.

(3) The digital twin is nowadays mainly used as an applied technology in the fields of robotics and autonomous driving to achieve relevant tasks directly by building digital twin entities or environments. However, the specific methodology and interpretation of related characteristics of digital twins are scarce in relevant literature. It is still a difficult task to achieve a comprehensive digital twin. With the help of meta-learning, the digital twin can obtain prior learning experience through the previous twin in the processes of environment interaction or simulated object synchronization, leading to the iterative optimization process.

(4) The models and algorithms of DT technologies are separated, and there is no comprehensive method to combine and evaluate them. For the model, environmental parameters, and some random noise in DTs, there is a lack of a synchronous sim2real method that can be considered as a dynamic process to improve the generalization of the transfer process.

(5) Parallel intelligence technologies improve autonomous driving systems' performance, handle big data, and solve the RG problem through multi-unit parallel computing and execution. With the expansion of the computing scale and increasement of functional requirements in the autonomous driving field, parallel intelligence technologies need to meet the requirements of ever-increasing computing scale and complexity.
	
(6) Parallel intelligence can help autonomous driving systems learn knowledge from the data generated by the physical and artificial systems, but the generated data exhibit characteristics of multi-modality, high massiveness and redundancy. Therefore, how to perform efficient data analysis to refrain from useless data and make a precise representation of the physical system are problems for PI-based autonomous driving systems.

%(7) There is a lack of system analysis of social sensing signals in many industries, such as how an operator or social attribute can influence the AVs in a parallel driving system. The medoling of social attribute in CPSS is quite challenging compared to physical attribute. The social attribute needs more system study and research.

\section{Conclusion}
Autonomous driving researches require a significant amount of real-world data to train a reliable and robust algorithm. On one hand, using real data in extreme and scarce scenarios often leads to a high cost, so simulated data are typically used to meet the requirement. On the other hand, there is always a gap between the simulated and real worlds, so it is necessary to further investigate methods for transferring knowledge from simulation to reality. In this paper, we comprehensively review state-of-the-arts of sim2real, DTs, and PI. In sim2real, we mainly introduce the methodologies and related applications from the view of different categories, while we focus on the virtual technologies about the DT dependency, and then introduce the common frameworks and applications of autonomous driving. Subsequently, the parallel intelligence thesis and technologies including parallel learning, parallel vision, and parallel driving, etc. in autonomous driving are reviewed. To demonstrate the simulation environment for realizing sim2real, DT and PI methods, we also summarize existing autonomous driving simulators. In addition, we present existing challenges for the future development of autonomous driving in sim2real, DTs and PI in this paper.

\bibliographystyle{elsarticle-num}
\bibliography{surveySim2realDTsParallel_v2}

\begin{IEEEbiography}[{\includegraphics[width=1in,height=1.25in,clip,keepaspectratio]{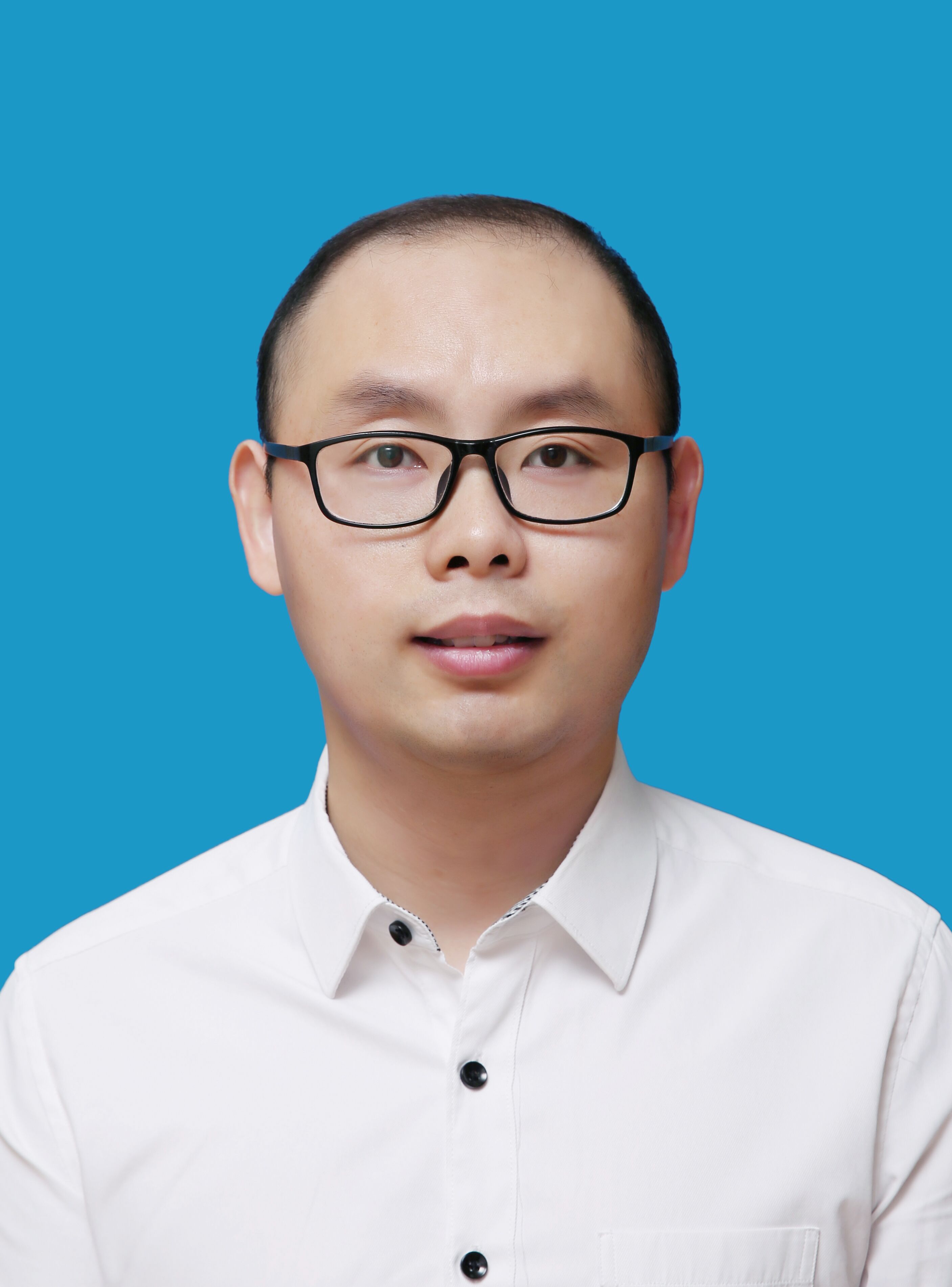}}]
	{Xuemin Hu}
	is currently an Associate Professor with School of Artificial Intelligence, Hubei University, Wuhan, China. He received the B.S. degree in Biomedical Engineering from Huazhong University of Science and Technology and the Ph.D. degree in Signal and Information Processing from Wuhan University in 2007 and in 2012, respectively. He was a visiting scholar in the University of Rhode Island, Kingston, RI, US from November 2015 to May 2016. His areas of interest include computer vision, machine learning, motion planning, and autonomous driving.In this paragraph you can place your educational, professional background and research and other interests.
\end{IEEEbiography}
\vspace{-10 mm}

\begin{IEEEbiography}[{\includegraphics[width=1in,height=1.25in,clip,keepaspectratio]{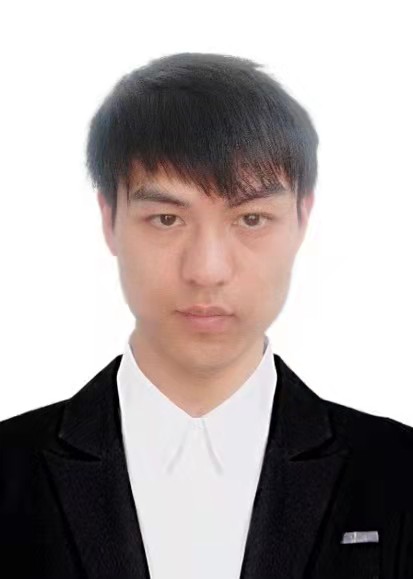}}]
	{Shen Li}
	received the B.S. degree in Computer Science and Technology from Liaoning University in 2018. From September 2022 to now, he is pursuing his Master's degree in School of Artificial Intelligence, Hubei University, Wuhan, China. His areas of interest include deep learning and autonomous driving.
\end{IEEEbiography}
\vspace{-10 mm}

\begin{IEEEbiography}[{\includegraphics[width=1in,height=1.25in,clip,keepaspectratio]{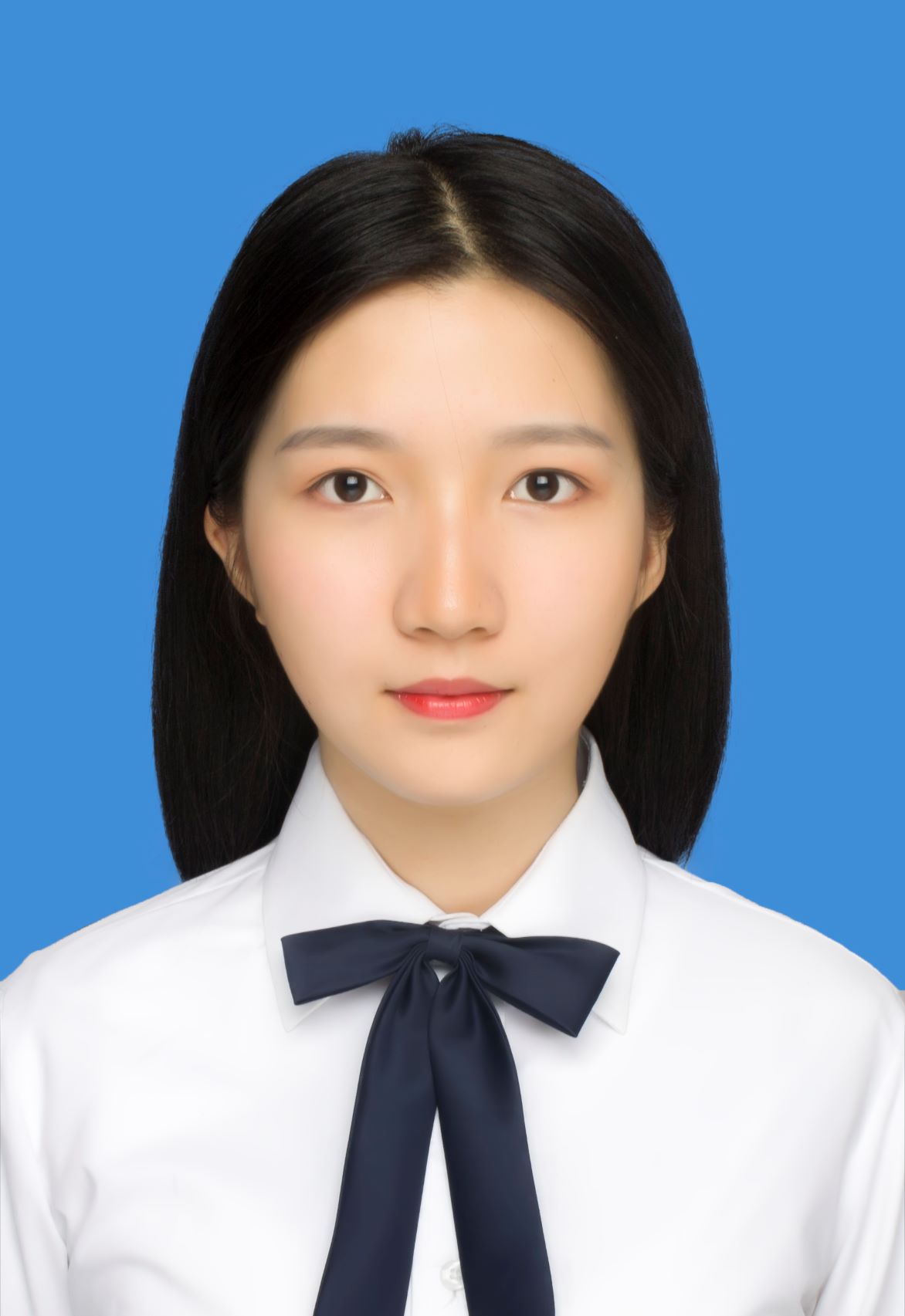}}]
	{Tingyu Huang}
	received the B.S. degree in Communication Engineering from Hunan Institute of Science and Technology in 2022. From September 2022 to now, she is pursuing her Master's degree in School of Artificial Intelligence, Hubei University, Wuhan, China. Her areas of interest include deep learning and autonomous driving.
\end{IEEEbiography}
\vspace{-10 mm}

\begin{IEEEbiography}[{\includegraphics[width=1in,height=1.25in,clip,keepaspectratio]{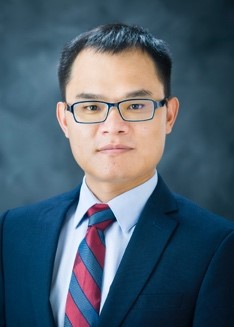}}]
	{Bo Tang}
	is an Associate Professor in the Department of Electrical and Computer Engineering at Worcester Polytechnic Institute. Prior to this, he was an Assistant Professor in the Department of Electrical and Computer Engineering at Mississippi State University. He received the Ph.D. degree in electrical engineering from University of Rhode Island (Kingstown, RI) in 2016. His research interests lie in the general areas of bio-inspired artificial intelligence (AI), AI security, edge AI, and their applications in Cyber-Physical Systems (e.g., wireless networks, autonomous vehicles, and power systems). He is currently an Associate Editor for IEEE Transactions on Neural Networks and Learning Systems.
\end{IEEEbiography}
\vspace{-10 mm}

\begin{IEEEbiography}[{\includegraphics[width=1in,height=1.25in,clip,keepaspectratio]{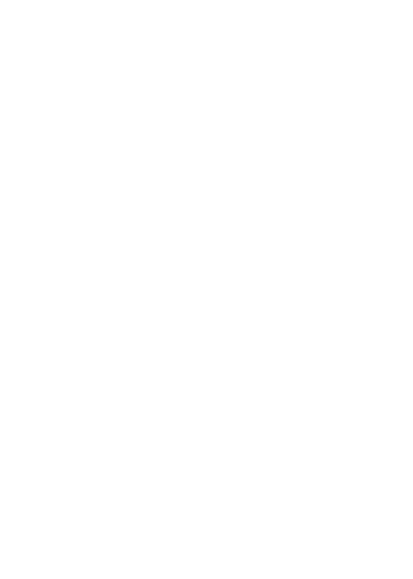}}]
	{Rouxing Huai}
currently an Advisory Scientist at Beijing Huairou Academy of Parallel Sensing, is a senior researcher in mechatronics, intelligent systems, color optics, and parallel optical fields. He received his PhD in Computer and Systems Engineering from USA in 1990 and worked at the University of Arizona for 21 years in Robotics and Intelligent Systems for Manufacturing, Space Exploration, and Optical Applications.
\end{IEEEbiography}
\vspace{-10 mm}

\begin{IEEEbiography}[{\includegraphics[width=1in,height=1.25in,clip,keepaspectratio]{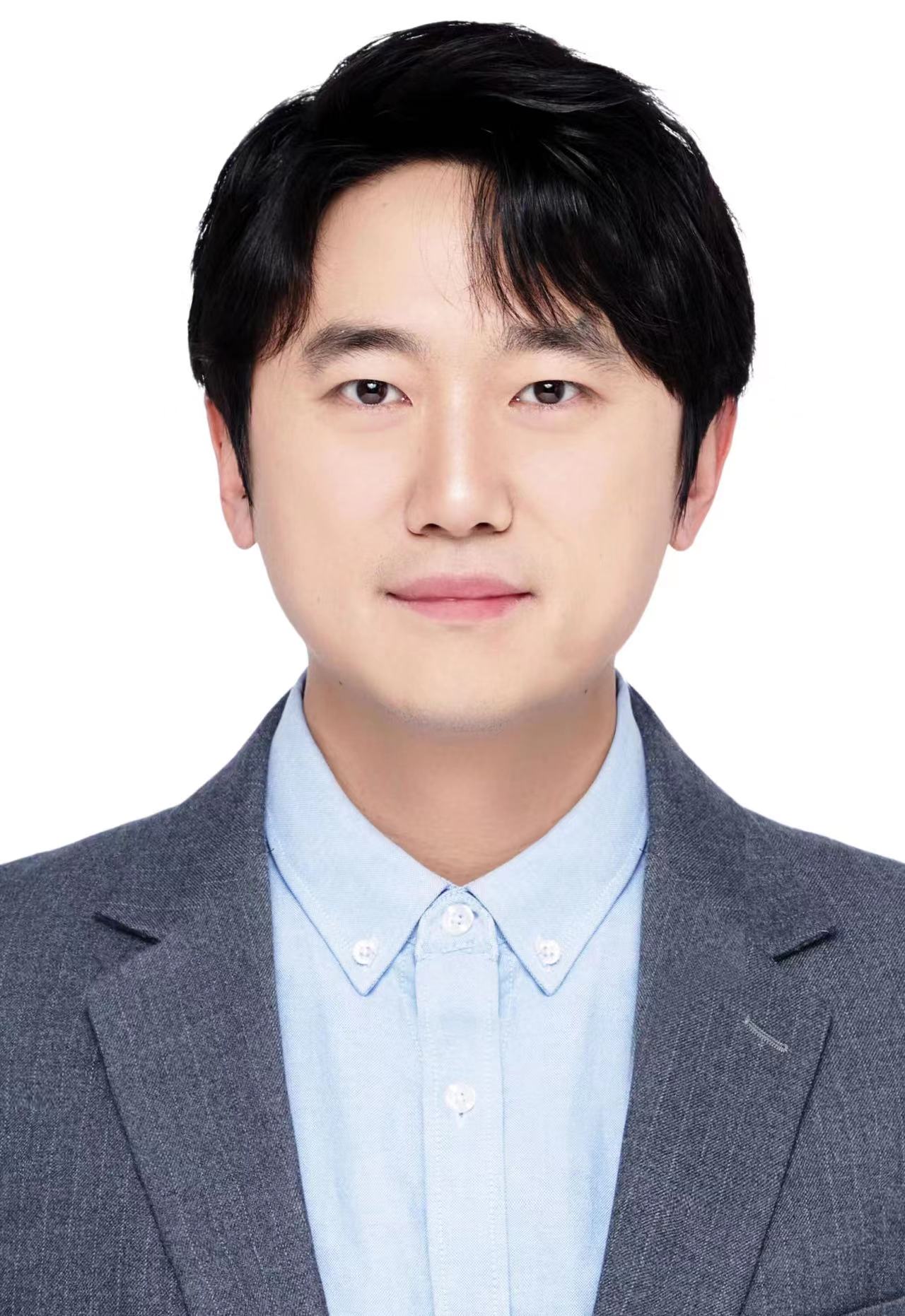}}]
	{Long Chen}
	is currently a Professor with State Key Laboratory of Management and Control for Complex Systems, Institute of Automation, Chinese Academy of Sciences, Beijing, China. His research interests include autonomous driving, robotics, and artificial intelligence, where he has contributed more than 100 publications. He received the IEEE Vehicular Technology Society 2018 Best Land Transportation Paper Award, the IEEE Intelligent Vehicle Symposium 2018 Best Student Paper Award and Best Workshop Paper Award, the IEEE Intelligent Transportation Systems Society 2021 Outstanding Application Award, the IEEE Conference on Digital Twin and Parallel Intelligence 2021 Best Paper and Outstanding Paper Award, the IEEE International Conference on Intelligent Transportation Systems 2021 Best Paper Award. He serves as an Associate Editor for the IEEE Transaction on Intelligent Transportation Systems, the IEEE/CAA Journal of Automatica Sinica, the IEEE Transaction on Intelligent Vehicle and the IEEE Technical Committee on Cyber-Physical Systems.
\end{IEEEbiography}

\vfill

\end{document}